\documentclass[lettersize,journal]{IEEEtran}
\usepackage{amsmath,amsfonts}
\usepackage{algorithmic}
\usepackage{algorithm}
\usepackage{array}
\usepackage[caption=false,font=normalsize,labelfont=sf,textfont=sf]{subfig}
\usepackage{textcomp}
\usepackage{stfloats}
\usepackage{url}
\usepackage{verbatim}
\usepackage{graphicx}
\usepackage{cite}
\usepackage{hyperref}
\begin{document}

\title{Non-Markov Multi-Round Conversational Image Generation with History-Conditioned MLLMs}

\author{Haochen Zhang, Animesh Sinha, Felix Juefei-Xu, Haoyu Ma, Kunpeng Li, Zhipeng Fan,\\ Meng Dong, Xiaoliang Dai, Tingbo Hou, Peizhao Zhang, Zecheng He}

\markboth{Journal of \LaTeX\ Class Files,~Vol.~14, No.~8, August~2021}%
{Shell \MakeLowercase{\textit{et al.}}: A Sample Article Using IEEEtran.cls for IEEE Journals}


\maketitle

\begin{abstract}
Conversational image generation requires a model to follow user instructions across multiple rounds of interaction, grounded in interleaved text and images that accumulate as chat history. While recent multimodal large language models (MLLMs) can generate and edit images, most existing multi-turn benchmarks and training recipes are effectively \emph{Markov}: the next output depends primarily on the most recent image, enabling shortcut solutions that ignore long-range history. In this work we formalize and target the more challenging \emph{non-Markov} setting, where a user may refer back to earlier states, undo changes, or reference entities introduced several rounds ago. We present (i) non-Markov multi-round data construction strategies, including rollback-style editing that forces retrieval of earlier visual states and name-based multi-round personalization that binds names to appearances across rounds; (ii) a history-conditioned training and inference framework with token-level caching to prevent multi-round identity drift; and (iii) enabling improvements for high-fidelity image reconstruction and editable personalization, including a reconstruction-based DiT detokenizer and a multi-stage fine-tuning curriculum. We demonstrate that explicitly training for non-Markov interactions yields substantial improvements in multi-round consistency and instruction compliance, while maintaining strong single-round editing and personalization.
\end{abstract}


\section{Introduction}
Recent progress in diffusion models~\cite{ho2020denoising, podell2023sdxl, rombach2022high} and multimodal large language models (MLLMs)~\cite{team2023gemini, ge2024seed, sun2024generative} has made it possible to generate high-quality images that follow detailed natural-language instructions. Beyond single-shot text-to-image synthesis, a growing set of applications---from creative design assistants to personalized visual storytelling and interactive content creation---requires \emph{conversational image generation}, where a user iteratively refines an image through multiple rounds of interaction. In this setting, the model must not only produce visually compelling outputs, but also track what has been established previously, resolve references in context, and apply new instructions in a way that remains consistent with both the conversation and the evolving image.

A key challenge is that multi-round image generation is fundamentally different from single-turn editing~\cite{brooks2023instructpix2pix, avrahami2022blended} or personalization~\cite{chen2024subject, wang2024instantid}. Real users interact in ways that are often \emph{non-linear}: they may ask to revert to an earlier version, undo a sequence of changes, or apply a new edit to a prior state rather than the most recent one. They may also introduce multiple entities early in the conversation and later refer to them by name or attributes that were defined several turns ago. Handling such behavior requires \textbf{non-Markov multi-round reasoning} over interleaved text and images, where the correct next image can depend on information that is \emph{not} contained in the most recent turn.

\begin{figure}[t]
\centering
\includegraphics[width=0.99\linewidth]{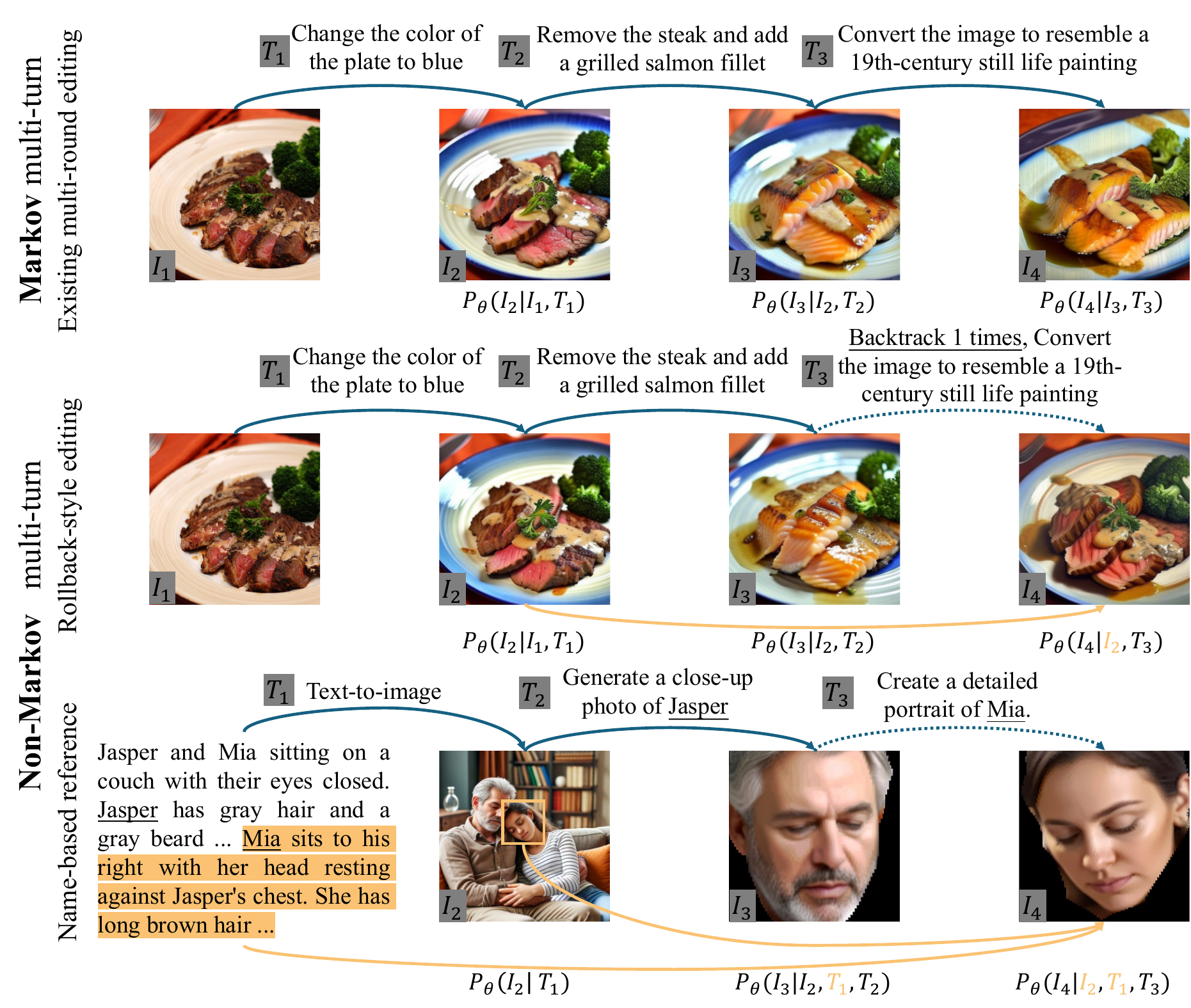}
\caption{\textbf{Non-Markov vs.\ Markov multi-round generation.} Markov: each turn depends primarily on the latest image. Non-Markov: later turns refer to earlier states (rollback/undo) or to entities introduced multiple rounds ago (name-based references).}
\label{fig:nonmarkov_overview}
\end{figure}

However, most existing multi-turn datasets~\cite{Zhang2023MagicBrush, ge2024seed} and training pipelines~\cite{sun2024generative, ge2024seed} implicitly simplify the problem to an \emph{effectively Markov} process, where the next output depends primarily on the latest image and the latest instruction. This Markov assumption can lead to shortcut learning: a model can appear strong in multi-turn settings while largely ignoring long-range context. In practice, this manifests as failures in rollback instructions (e.g., ``go back to the version before the background change''), inconsistent identity for personalized subjects across turns, and incorrect entity references when multiple people or objects are involved. These failure modes motivate a more faithful formulation of conversational image generation that explicitly targets the non-Markov setting.

In this work, we advocate for and develop a solution to \textbf{non-Markov multi-round conversational image generation}. We formalize the distinction between Markov and non-Markov multi-round generation and show why the latter better reflects realistic user behavior. We then introduce practical data construction strategies to create non-Markov supervision at scale. First, we construct \textbf{rollback-style multi-round editing} examples by augmenting Markov editing chains with edits originating from earlier states and composing them into dialogs that explicitly refer back to those earlier versions. These dialogs require the model to retrieve the correct historical visual state before applying a new instruction, making them inherently non-Markov. Second, we introduce \textbf{name-based multi-round personalization} examples in which a subject is introduced and named in an early round, and later rounds request portraits or refinements using only the name. We construct this data from videos, using captioning and name assignment to define entities and face detection/segmentation to extract supervised targets, thereby enforcing persistent name-to-appearance binding across rounds.

Solving non-Markov multi-round generation also raises systems-level and modeling challenges. A practical obstacle is \textbf{multi-round drift}: repeatedly encoding a generated image into tokens and decoding back to pixels can introduce small reconstruction errors that accumulate over turns, leading to identity degradation and compounding artifacts. To address this, we propose \textbf{token-level history caching}, where the model caches and reuses previously generated image tokens directly as dialog history, rather than re-encoding reconstructed images at each round. This design reduces compounding reconstruction noise and stabilizes long-horizon interactions.

Finally, robust non-Markov multi-round performance depends on strong underlying capabilities in reconstruction fidelity and personalization. We therefore incorporate two enabling improvements. We upgrade the diffusion-based detokenizer with a \textbf{reconstruction-based DiT detokenizer}~\cite{peebles2023scalable} that preserves fine details (especially faces) while maintaining compatibility with the tokenizer interface, minimizing the need for upstream retraining. In addition, we adopt a \textbf{multi-stage fine-tuning curriculum} for editable personalization that transitions from identity-preserving reconstruction to prompt-following edits while maintaining subject consistency. Together, these components form a coherent framework: high-fidelity reconstruction and editable personalization provide a reliable substrate, while non-Markov data construction and history-aware training/inference unlock true conversational reasoning beyond the Markov assumption.

Our contributions are summarized as follows:
\begin{itemize}
    \item \textbf{Non-Markov formulation for multi-round conversational image generation.} We formalize the limitation of Markov-style multi-turn settings and motivate non-Markov multi-round reasoning as the appropriate target for realistic conversational behavior.
    \item \textbf{Non-Markov multi-round data construction.} We propose rollback-style editing construction from existing Markov dataset, and name-based multi-round personalization from videos, producing supervision that requires retrieving earlier visual states and long-range entity bindings.
    \item \textbf{History-conditioned training and token-level caching.} We train with full dialog history and introduce token-level caching to mitigate multi-round drift caused by repeated encode/decode cycles.
    \item \textbf{Enabling improvements.} We introduce a reconstruction-based DiT detokenizer and a multi-stage personalization curriculum that together improve fidelity and editable identity preservation, which are critical for stable multi-round interactions.
\end{itemize}

This manuscript is an extended version of our published conference paper~\cite{ourwacv}, with a primary focus on non-Markov multi-round conversational image generation, including additional datasets, detailed methods, and more evaluations. The rest of the paper is organized as follows. Section~\ref{sec:formulation_data} formalizes Markov and non-Markov multi-round generation and presents our non-Markov data construction pipelines. Section~\ref{sec:method} describes history-conditioned training, token-level caching, and key enabling components. Section~\ref{sec:experiments} reports experimental results, emphasizing non-Markov rollback editing and name-based multi-round personalization, followed by ablations and analysis.


\section{Related Work}
\subsection{Image Generation, Editing, and Personalization}
Diffusion models, beginning with DDPM~\cite{ho2020denoising}, have rapidly advanced text-to-image generation, with systems such as Stable Diffusion~\cite{podell2023sdxl, rombach2022high}, DALL-E~\cite{ramesh2022hierarchical}, and Imagen~\cite{baldridge2024imagen} producing high-fidelity images via iterative denoising. To strengthen \emph{visually conditioned} generation, control mechanisms such as ControlNet~\cite{zhang2023adding} and T2I-Adapter~\cite{mou2024t2i} have been widely adopted, enabling applications including image editing~\cite{brooks2023instructpix2pix, avrahami2022blended}, composition~\cite{wang2024instantid}, subject-driven generation~\cite{chen2024subject}.

In this paper, we focus on two closely related conditional generation regimes: \textbf{instruction-guided image editing} and \textbf{human-centric subject-driven generation} (personalization for short). For editing, approaches such as InstructPix2Pix~\cite{brooks2023instructpix2pix} and HIVE~\cite{zhang2024hive} enable instruction-following edits, while methods such as Imagic~\cite{kawar2023imagic} and Text2LIVE~\cite{bar2022text2live} offer fine-grained control over local and global appearance. For personalization, early methods including Textual Inversion~\cite{gal2022image} and DreamBooth~\cite{ruiz2023dreambooth} adapt diffusion models to a specific identity via fine-tuning, but often suffer from limited generalization and user-specific optimization overhead. Recent works aim to improve efficiency and generality by fusing visual and textual cues, for example by mapping vision features into the text embedding space (e.g., ELITE~\cite{wei2023elite}) or fusing vision and text tokens via cross-attention (e.g., PhotoMaker~\cite{li2024photomaker}), and more recent unified diffusion systems (e.g., OmniGen~\cite{xiao2024omnigen}) attempt to consolidate editing and personalization capabilities in a single model.

Beyond diffusion, autoregressive generators have also been explored for image synthesis (e.g., LlamaGen~\cite{sun2024autoregressive} and Fluid~\cite{fan2024fluid}), offering an alternative modeling paradigm. Meanwhile, some modern MLLMs, such as SEED-X~\cite{ge2024seed} and EMU2~\cite{sun2024generative}, have demonstrated promising image generation and editing capabilities, suggesting that a unified token-based interface can enable both multimodal reasoning and visual synthesis within a single framework.

\subsection{Multimodal Large Language Models for Vision-Language Reasoning and Generation}
Large language models~\cite{openai2023chatgpt, openai2023gpt4, touvron2023llama} have shown impressive performance in language generation and reasoning, motivating extensions that incorporate visual inputs for vision-language understanding~\cite{liu2024visual, peng2023kosmos, chen2023minigpt, zhu2023minigpt}. Vision-language models such as LLaVA~\cite{liu2023llava} integrate visual perception into LLMs through lightweight vision-language connectors and instruction data, enabling strong image-question answering and multimodal reasoning.

More recently, generative MLLMs~\cite{team2023gemini, ge2024seed, sun2024generative} have emerged that can produce both text and images, treating images as a ``new language'' via a visual tokenizer/detokenizer pair. Representative directions include models that generate discrete image tokens (e.g., Chameleon-style~\cite{team2024chameleon} architectures) and models that operate in continuous feature spaces (e.g., CLIP-feature-based tokenization~\cite{bai2023qwen, sun2023eva} as in EMU serise~\cite{sun2023generative, sun2024generative} and SEED-X~\cite{ge2024seed} systems). These generative MLLMs are appealing for conversational image generation: they can, in principle, reason over interleaved text and images and produce images as part of the response. However, their performance in conditional image-to-image generation (e.g., editing and personalization) depends critically on the visual tokenization and, in particular, the detokenizer's ability to preserve fine-grained content such as human identity. Despite demonstrating strong capabilities in interleaved text–image generation, to the best of our knowledge, conversational image generation has not yet been explored.

\subsection{Multi-Round Image Generation Datasets}
Training a conversational image generation model requires multi-round instruction data with \emph{text--image interleaved} histories. Many popular multi-round vision-language datasets (e.g., LLaVA-style~\cite{liu2023llava} instruction data, SVIT~\cite{zhao2023svit}, and related variants~\cite{zhang2023llavar}) are primarily designed for \emph{text outputs} and do not directly supervise image generation. Some interleaved corpora (e.g., MMC4~\cite{zhu2024multimodal}, VIST~\cite{huang2016visual}, and related instruction datasets\cite{xu2024lateralization}) can be adapted into multi-round text--image sequences, but the images are often not temporally or semantically consistent across turns, limiting their utility for conditional generation tasks such as editing and personalization. For instance, the VIST dataset comprises five images and their corresponding captions to narrate a short story, but the images may show inconsistent elements, such as two players in one image and only a football in the next, making it challenging to train models for editing or personalization tasks.

Datasets most aligned with our needs are multi-turn editing datasets that provide a source image and a sequence of editing prompts, such as MagicBrush~\cite{Zhang2023MagicBrush} and SEED-Data-Edit~\cite{ge2024seed}. However, these datasets typically follow an \textbf{Markov property}: each target image is an edited version of the previous target, which substantially reduces reasoning requirements in the multi-turn setting because the model can learn a shortcut by relying primarily on the immediate previous output. In contrast, real user conversations are often \textbf{non-Markov}: later turns may refer back to earlier states (rollback/undo) or to entities introduced several rounds ago (e.g., name-based references). This gap motivates our construction of non-Markov multi-round datasets and our history-conditioned training/inference design for conversational multi-round image generation.


\section{Problem Formulation and Non-Markov Dataset Construction}
\label{sec:formulation_data}

Conversational image generation proceeds over multiple rounds where each user instruction should be interpreted in the context of an interleaved text--image chat history. A core difficulty is that the correct next image may depend on information introduced several turns earlier. In this section, we first clarify the difference between \emph{Markov} and \emph{non-Markov} multi-round generation, and then describe how we build two non-Markov datasets that explicitly require long-range history reasoning.

\subsection{Markov vs.\ Non-Markov Multi-Round Generation}
We represent a conversation up to turn $t$ as an interleaved history
\begin{equation}\label{eq:1}
\mathcal{H}_t = \{(T_1, I_1), (T_2, I_2), \dots, (T_t, I_t)\},
\end{equation}
where $T_i$ is the user instruction at round $i$ and $I_i$ is the model-generated image at that round. Here, we focus on image outputs for simplicity. Given $\mathcal{H}_t$ and a new instruction $T_{t+1}$, the model generates the next image $I_{t+1}$:
\begin{equation}\label{eq:2}
I_{t+1} \sim p_\theta(I \mid T_{t+1}, \mathcal{H}_t).
\end{equation}

\subsubsection{Markov multi-round generation}
Many existing multi-turn editing benchmarks~\cite{Zhang2023MagicBrush, ge2024seed} are \emph{effectively Markov}: each round is an edit of the immediately previous result. In such a setting, the next output can be approximated using only the latest image:
\begin{equation}
p_\theta(I_{t+1} \mid T_{t+1}, \mathcal{H}_t) \approx p_\theta(I_{t+1} \mid T_{t+1}, I_t).
\end{equation}
This encourages a shortcut solution where a model can ignore long-range history and still perform well, because the most recent image already contains almost all useful information.

\subsubsection{Non-Markov multi-round generation}
Real user interactions are often \emph{non-Markov}: later instructions may refer back to an earlier state (e.g., ``undo the last two changes''), or to an entity introduced several rounds ago (e.g., ``generate a close-up of $\{name\}$'' where the name--appearance binding was established earlier). In these cases, conditioning only on $I_t$ is insufficient, and success requires retrieving and grounding the instruction in earlier parts of the chat history (\emph{both text and images}). Non-Markov behavior is therefore a more faithful target for conversational image generation, but is underrepresented in existing training data.

\subsection{Dataset I: Rollback-Style Non-Markov Multi-Round Editing}
\label{sec:rollback_data}
We construct a non-Markov multi-round editing dataset that explicitly requires \textbf{state selection} from history, as depicted in Fig.~\ref{fig:rollback}. We start from existing \emph{Markov} editing chains that provide a sequence of edits applied to an initial image:
\[
A_1 \xrightarrow{T_2} A_2 \xrightarrow{T_3} A_3 \xrightarrow{T_4} A_4 \xrightarrow{T_5} A_5,
\]
where $A_{i+1}$ is the edited image after round $i$ and $T_{i+1}$ is the corresponding edit instruction.

\subsubsection{Step 1: create additional edits from earlier states}
For an earlier state (e.g., $A_1$ or $A_2$), we sample additional single-round editing instructions and generate corresponding targets:
\[
A_1 \xrightarrow{T_i} B_i (i\ge 3),\quad A_2 \xrightarrow{T_j} C_j (j\ge 4),\ \dots
\]
These extra branches provide valid alternative outcomes originating from earlier images.

\subsubsection{Step 2: synthesize rollback dialogs}
We then compose a \textbf{non-Markov} final instruction that explicitly refers back to an earlier state, e.g.,
\begin{quote}
``Undo the last two changes, and \textit{[apply a new edit]}.''
\end{quote}
or
\begin{quote}
``Instead of \textit{[previous edit]}, \textit{[apply a new edit]}.''
\end{quote}
The supervision target is the image produced by applying the new edit to the referenced earlier state (e.g., $A_3$ for ``undo the last two changes''), not to the most recent image $A_5$. As a result, a Markov shortcut that relies only on $A_5$ will fail, while a history-aware model that can retrieve the correct earlier state can succeed.

\subsubsection{Why this dataset is non-Markov.}
The final instruction cannot be reliably solved using the latest image alone; correctness requires using the full history to determine which prior state is the intended base. However, this task is relatively simple as the MLLMs are just required to learn to pick up a related $I_{base}$ from chat history $\mathcal{H}_t$:  

\begin{equation}
p_\theta(I_{t+1} \mid T_{t+1}, \mathcal{H}_t) \approx p_\theta(I_{t+1} \mid T_{t+1}, I_{base}).
\end{equation}

Therefore, we further design a name-based multi-round personalization task which requires token-level reasoning from both textual inputs and visual outputs.

\begin{figure}[t]
\centering
\includegraphics[width=0.99\linewidth]{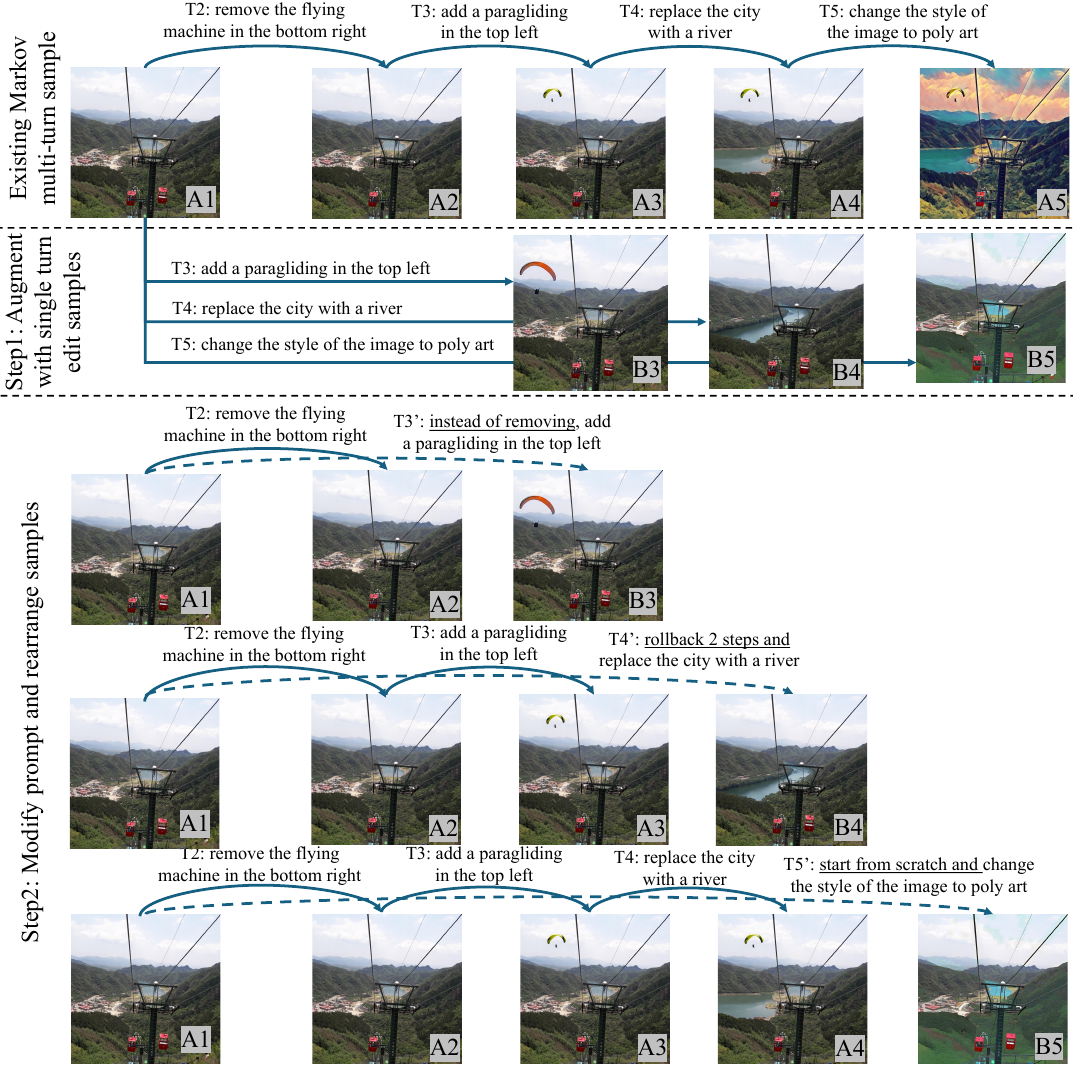}
\caption{\textbf{Rollback-style non-Markov editing construction.} Starting from Markov editing chains, we add alternative edits from earlier states and synthesize final-round rollback instructions so the correct target depends on an earlier image rather than the latest one.}
\label{fig:rollback}
\end{figure}

\subsection{Dataset II: Name-Based Non-Markov Multi-Round Personalization}
\label{sec:name_data}
In this section, we build a non-Markov dataset that requires \textbf{entity binding} across turns. The key idea is to introduce multiple people in an early round, assign them names, and later request images using only the names. This tests whether the model can bind a symbolic reference (name) in formate of text to a visual identity (face) established earlier.

\begin{figure}[t]
\centering
\includegraphics[width=0.99\linewidth]{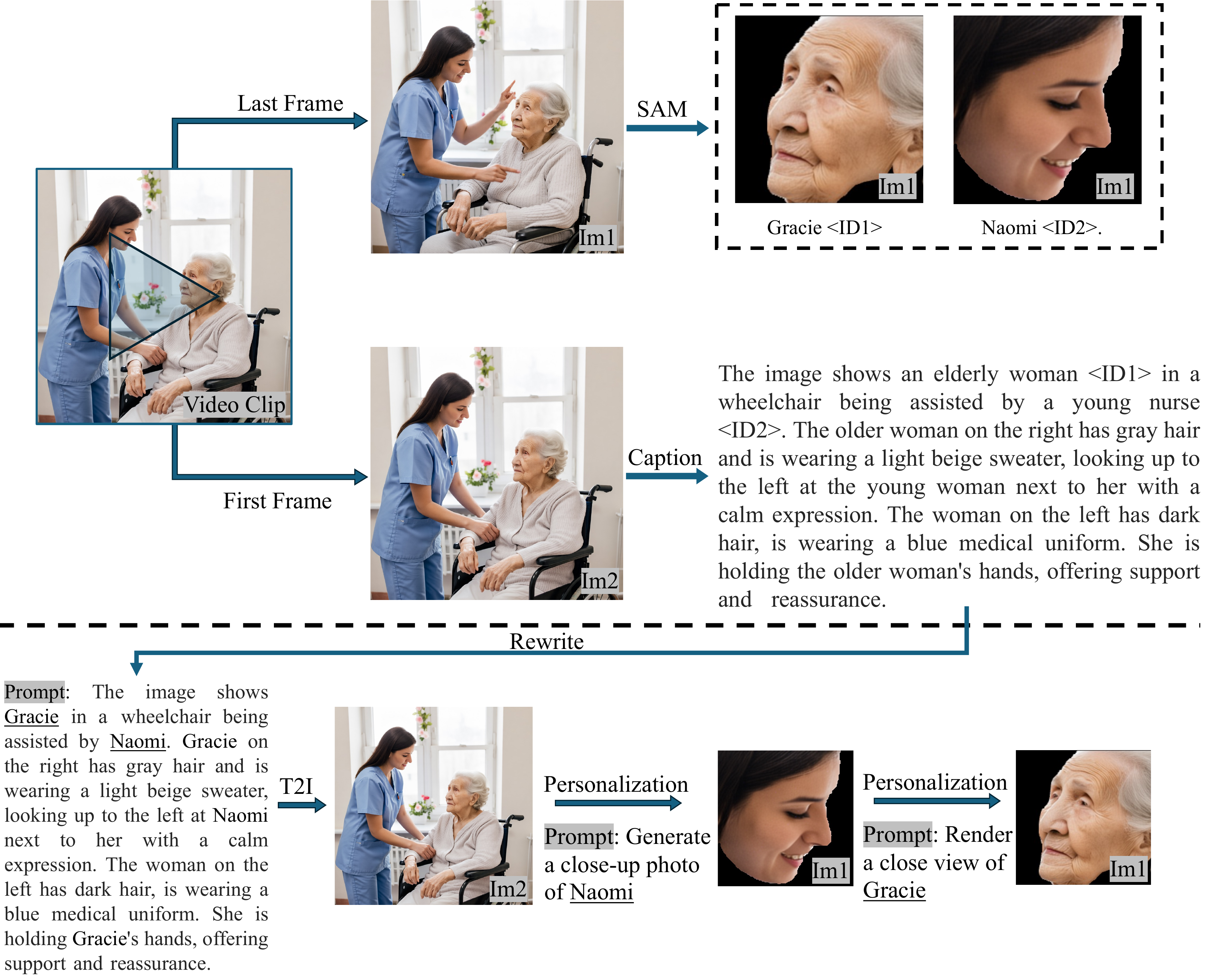}
\caption{\textbf{Name-based non-Markov multi-round personalization from video.} Round 1 introduces and names multiple people. Later rounds request portraits by name only. Supervision comes from video-derived identity targets, enforcing persistent name$\leftrightarrow$appearance binding across rounds.}
\label{fig:name_dataset}
\end{figure}

\subsubsection{Conversation Structure.}
We design a three-round conversational interaction as follows:
\begin{itemize}
    \item \textbf{Round 1 (T2I):} generate an image containing multiple people described in text, where each individual is explicitly assigned a unique name (e.g., ``Olivia'' and ``Julian'').
    \item \textbf{Round 2 (short-term reference):} request a close-up portrait of one person using only their name (e.g., ``Generate a portrait of Olivia'').
    \item \textbf{Round 3 (long-term reference):} request a close-up portrait of a different person using only their name (e.g., ``Generate a portrait of Julian'').
\end{itemize}

Ideally, both the second and third rounds would generate full-body or full-scene personalized images rather than face-centric portraits. However, constructing real-world supervision for such multi-person, multi-round personalization is challenging, as it is difficult to obtain aligned triplets consisting of (i) an image of person\_1, (ii) an image of person\_2, and (iii) an image jointly containing both individuals. In practice, we therefore build this dataset from video clips featuring two people, which naturally provide multiple views of each individual while preserving identity consistency across frames. This design allows us to isolate and evaluate long-range name-to-identity binding in a controlled yet realistic non-Markov setting. However, we provide an exploration of \textbf{full-body} personalization in Sec.~\ref{sec:exp_fullbody}.

\subsubsection{Video-Based Supervision Pipeline.}
We construct supervision from videos to obtain consistent identity targets across multiple rounds, as illustrated in Fig.~\ref{fig:name_dataset}. The pipeline proceeds as follows:
\begin{itemize}
    \item Sample a video frame $F_1$ and generate a caption that describes multiple individuals present in the scene.
    \item Use a large language model~\cite{dubey2024llama} to assign unique names to each described individual and rewrite the caption to explicitly include these name--entity associations.
    \item Sample another frame $F_2$ from the same video; apply face detection (e.g., ArcFace~\cite{deng2019arcface}) and segmentation (e.g., SAM~\cite{kirillov2023segment}) to extract a clean face crop corresponding to the target individual, which serves as supervision for name-referenced portrait generation in later rounds.
\end{itemize}
This pipeline produces training examples in which the model must retrieve the name--appearance binding established in the first round, grounded jointly in text and images, and generate the correct identity in subsequent rounds using only the symbolic name reference.

\begin{figure*}[t]
  \centering
   \includegraphics[width=0.99\linewidth]{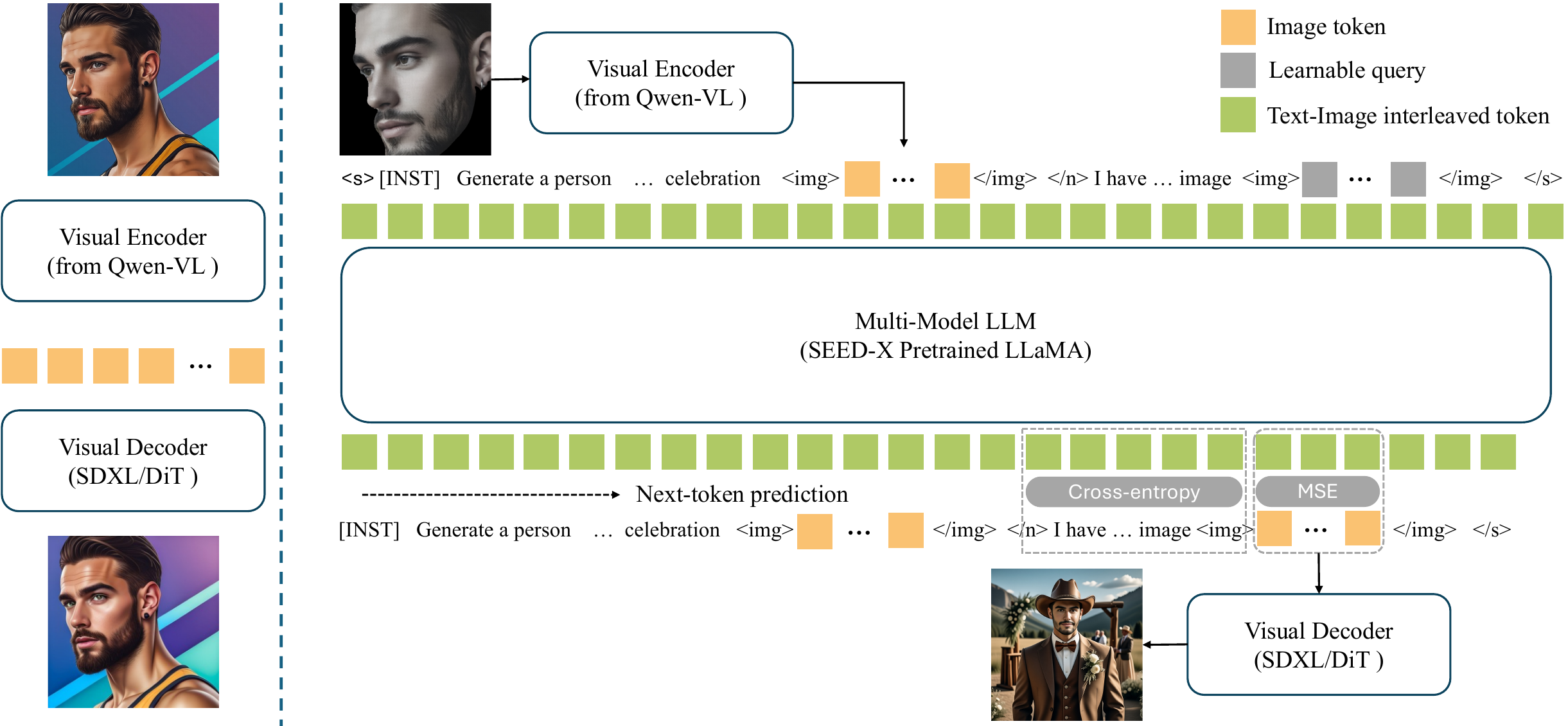}
   \caption{\textbf{Overview of our framework and data pipeline.}}
   \label{fig:seedx_pipe}
\end{figure*}

\subsubsection{Why This Dataset Is Non-Markov}
In later rounds, user prompts may contain only a name, without restating any visual attributes. Crucially, the third-round request refers to an identity that was introduced in the first round but is \emph{not} present in the immediately preceding second round output. Conditioning generation solely on the most recent image therefore provides insufficient information to resolve the requested identity.

Formally, while a Markov assumption would approximate generation as
\[
p_\theta(I_3 \mid T_3, \mathcal{H}_2) \approx p_\theta(I_3 \mid T_3, I_2),
\]
the correct dependency in this setting is closer to
\[
p_\theta(I_3 \mid T_3, \mathcal{H}_2) \approx p_\theta(I_3 \mid T_3, T_1^{(e)}, I_1^{(e)}),
\]
where $T_1^{(e)}$ and $I_1^{(e)}$ denote the portions of the first-round text and image that define the appearance of the referenced entity $e$. Correct generation therefore requires reasoning over earlier conversational context and selectively retrieving the appropriate name--appearance binding, rather than relying on the immediately preceding output. This makes the dataset inherently non-Markov and qualitatively more challenging than rollback-style multi-round editing, which typically involves selecting among prior visual states rather than resolving token-level symbolic-to-visual bindings.


\section{Method}
\label{sec:method}

This section describes our history-conditioned training and inference framework for non-Markov multi-round conversational image generation, followed by two key enabling components: (i) a reconstruction-based DiT detokenizer for high-fidelity image decoding, and (ii) a multi-stage instruction fine-tuning strategy for editable personalization.

\subsection{History-Conditioned Training and Token-Level Caching}
\label{sec:hist_train_cache}

Our framework is built upon a generative MLLM where images are represented as tokens and interleaved with text tokens.  In detail, we adopt Fig.~\ref{fig:seedx_pipe}, a SEED-X style architecture: a frozen vision encoder tokenizes an image $I$ into a sequence of visual tokens, and an autoregressive LLM predicts the next tokens in an interleaved text--image sequence. In our implementation, the vision encoder produces a fixed-length sequence of visual tokens (e.g., pooled to 64 tokens) that serves as the ``image language'' consumed and predicted by the LLM.

\subsubsection{History-conditioned input sequence}
For multi-round generation, we concatenate the entire chat history and the current instruction into one interleaved sequence:
\begin{equation}\label{eq:5}
S_{t+1} = [T_1, \Phi(I_1), \dots, T_t, \Phi(I_t), T_{t+1}]
\end{equation}
where $\Phi(\cdot)$ denotes the image tokenization process (vision encoder + pooling). The model then generates the image-token sequence $\hat{I}_{t+1}$, which is decoded into pixels $I_{t+1}$ by a detokenizer (Sec.~\ref{sec:dit_detok}).

\subsubsection{Final-Round Image Supervision}
\label{sec:final_round_loss}
In principle, one could supervise image-token predictions at \emph{every} round in a multi-turn sequence like LLaVA~\cite{liu2023llava}. In practice, however, our design is constrained by the SEED-X generation interface. Specifically, to generate an image token sequence, SEED-X conditions the LLM on a fixed set of 64 learnable query tokens that serve as the starting queries for image-token prediction. The predicted image tokens are thus generated \emph{from these learnable queries} conditioned on the current interleaved context.

This design introduces an important limitation for multi-image supervision within a single training sample: if we attempt to predict multiple images sequentially in one forward pass, later image predictions are not naturally conditioned on the earlier predicted image tokens as informative context. Instead, the later image tokens are again generated from the same learnable query tokens, which can decouple the later predictions from the earlier visual states and lead to an undesirable learning signal for multi-round behavior.

To bypass this limitation while still leveraging long histories, we split each multi-round training example into multiple prefixes, and supervise only the final-round image tokens in each prefix. Concretely, given an interleaved multi-round sequence
\[
A_1 \xrightarrow{T_2} A_2 \xrightarrow{T_3} A_3 \xrightarrow{T_4} A_4,
\]
we construct multiple training instances corresponding to different history lengths:
$(A_1, T_2 \rightarrow A_2)$,$(A_1, T_2, A_2, T_3 \rightarrow A_3)$,$(A_1, T_2, A_2, T_3, A_3, T_4 \rightarrow A_4)$ where each instance uses the full available prefix as context and computes image-token loss only on the target image at the final round of that prefix. This ensures that the model is always trained to predict a single image token sequence per sample under the SEED-X interface, while still exposing the model to varying-length histories and multi-round conversational dependencies.

For each constructed prefix instance, we apply cross-entropy loss on text tokens as usual and apply MSE loss only on the final target image tokens.

\begin{figure}[t]
  \centering
   \includegraphics[width=0.9\linewidth]{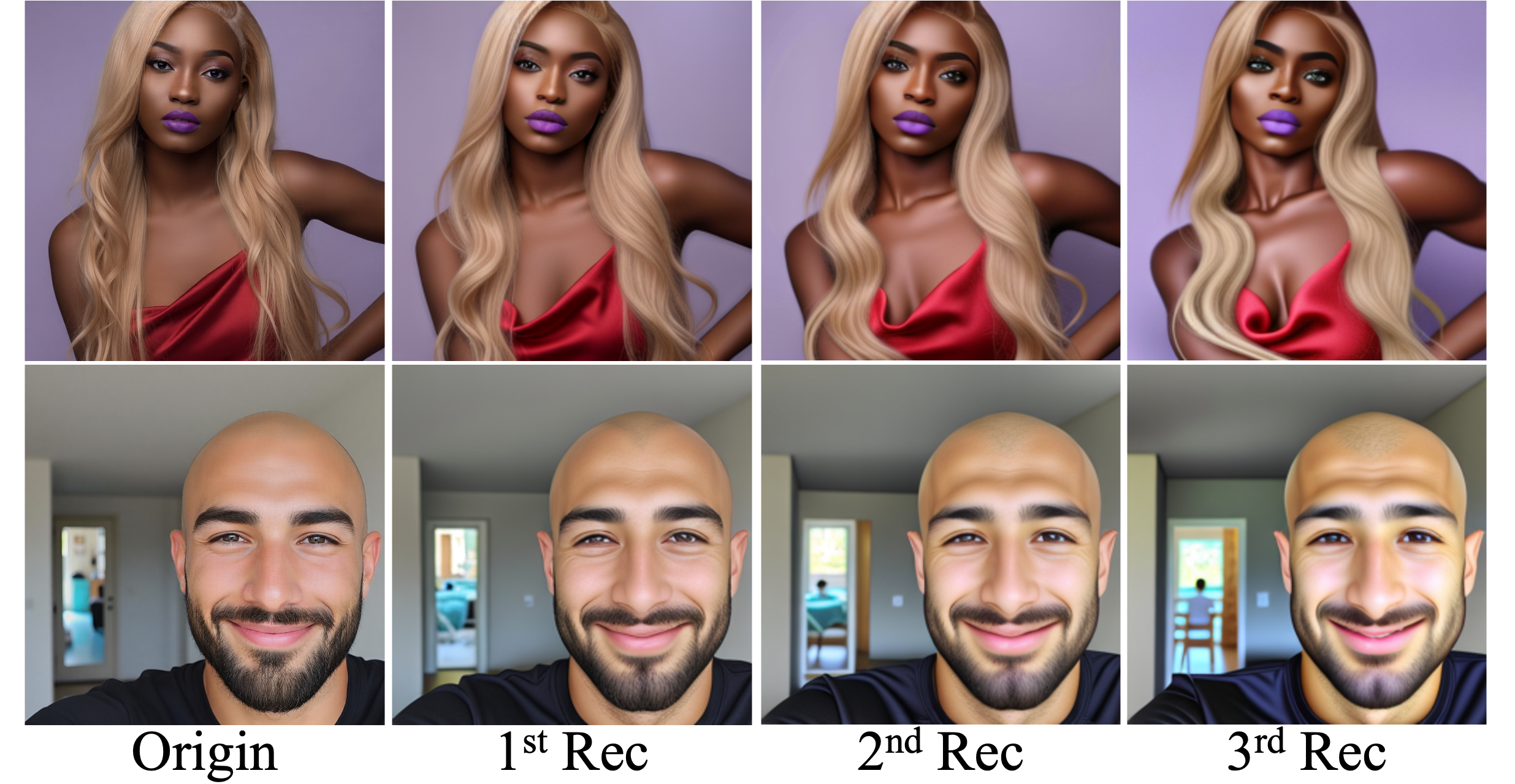}
   \caption{\textbf{Illustration of accumulated error when encoding and decoding an image several times.} This result suggests to caching image tokens in chat history instead of image pixels when performing multi-round inference.}
   \label{fig:detokenizer_error}
\end{figure}

\subsubsection{Token-level history caching for inference.}
A practical failure mode in multi-round interactions is \textbf{drift}: if each generated image $I_t$ is repeatedly re-encoded into tokens and then decoded back to pixels for the next turn, small reconstruction errors accumulate and gradually corrupt identity and content. Please refer to Fig.~\ref{fig:detokenizer_error} for examples. 

To mitigate this, we introduce \textbf{token-level caching} for inference. At round $t$, the model produces image tokens $\hat{I}_t$ (before detokenization). We cache $\hat{I}_t$ and reuse it directly as the history representation for later rounds: $\Phi(I_t)\ \leftarrow\ \hat{I}_t$ (cached tokens instead of re-encoding). Thus Eq~(\ref{eq:5}) changes to 

\begin{equation}\label{}
S_{t+1} = [T_1, \hat{I}_1, \dots, T_t, \hat{I}_t, T_{t+1}]
\end{equation}

This avoids repeated encode--decode cycles and substantially improves multi-round stability, which is especially important for non-Markov interactions where the model must reliably retrieve earlier identities or states over long horizons.


\subsection{Enabling Component I: Reconstruction-Based DiT Detokenizer}
\label{sec:dit_detok}

\begin{figure}[t]
  \centering
   \includegraphics[width=0.99\linewidth]{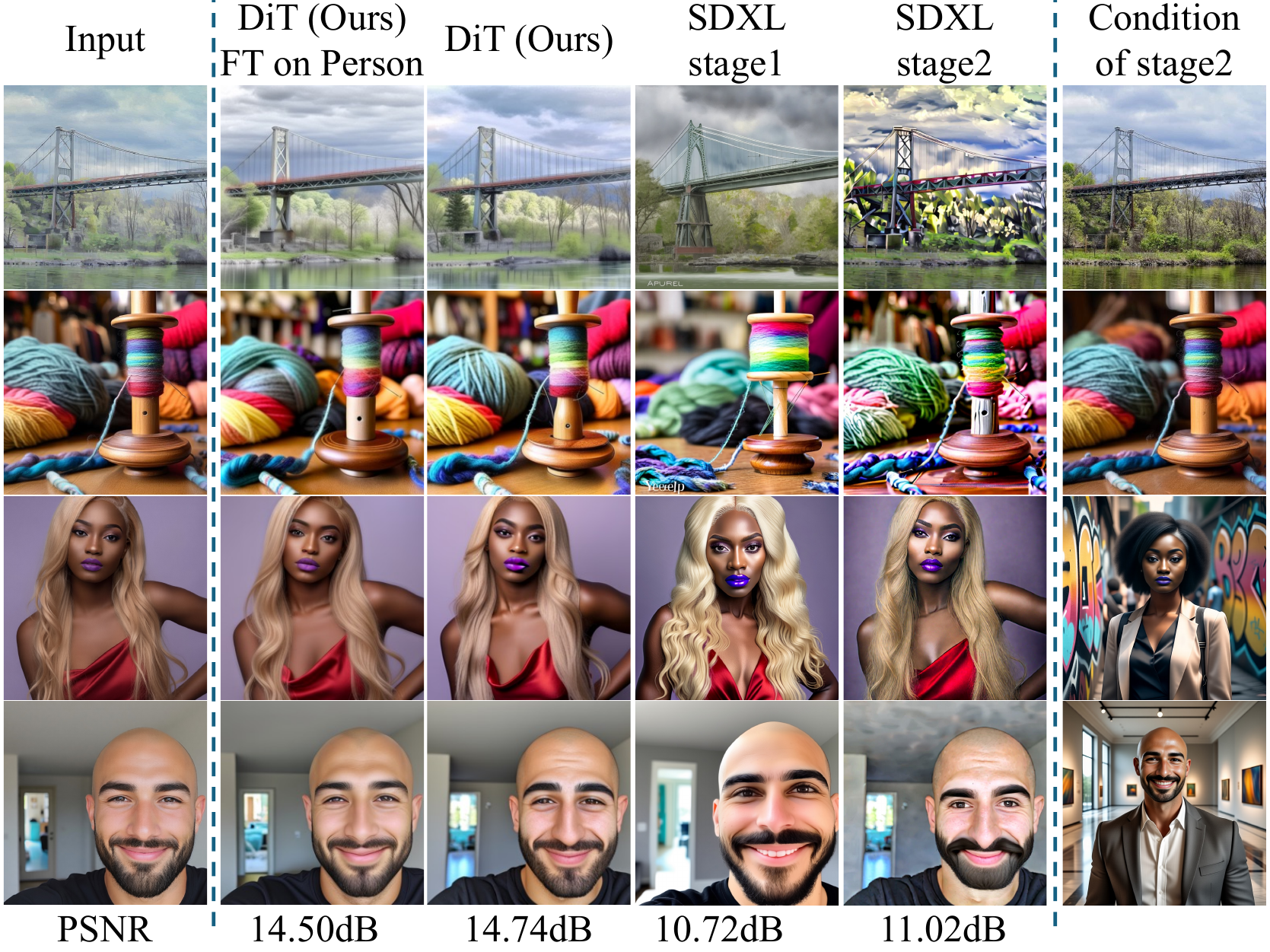}
   \caption{\textbf{Detokenizer performance comparison.} SEED-X stage1 detokenizer struggles with detail preservation. Stage2 uses a condition image to keep shapes but introduces artifacts and fails to maintain faces. Our DiT detokenizer preserves details without additional conditions and reconstructs faces well after tuning on human images.}
   \label{fig:detok_compare}
\end{figure}

High-quality detokenization is critical for conversational personalization: the model must preserve fine-grained identity cues (e.g., facial geometry, skin texture) across turns. We identify detokenization as a major bottleneck in existing generative MLLMs~\cite{ge2024seed}, particularly for human faces.

\subsubsection{Limitations of prior detokenizers.}
A common approach (e.g., SEED-X style~\cite{ge2023making}) uses an SDXL-based diffusion detokenizer~\cite{podell2023sdxl} with two stages:
\begin{itemize}
    \item \textbf{Stage-1 reconstruction detokenizer:} Conditioned on the image tokens, the diffusion decoder reconstructs the input image. While general, it often fails to preserve fine details, especially facial identity like in Fig.~\ref{fig:detok_compare}.
    \item \textbf{Stage-2 editing-tuned detokenizer:} To improve structure preservation, a conditional image is concatenated to the diffusion noise map and the decoder is fine-tuned on editing data. This can work well when the conditional image and target are in an \emph{editing relationship}, but it often generalizes poorly outside that regime (e.g., identity preservation), and can introduce artifacts when the condition image is out-of-distribution.
\end{itemize}

\subsubsection{Our DiT detokenizer.}
To improve both reconstruction fidelity and generalizability, we replace the diffusion detokenizer with a more powerful reconstruction-based Diffusion Transformer (DiT)~\cite{peebles2023scalable} detokenizer. Importantly, we retain the same image token interface (same frozen vision encoder and token pooling) so that the LLM component does not require costly re-pretraining.

We initialize the DiT detokenizer from a text-to-image diffusion model~\cite{polyak2024movie} and fine-tune it for reconstruction, analogous in spirit to Stage-1 reconstruction training but with improved capacity and inductive bias for token-conditioned decoding. Since personalization is especially sensitive to facial details, we further fine-tune the DiT detokenizer on human images under the same reconstruction objective, which empirically improves identity fidelity without introducing conditional-image dependencies. Thus our detokenizer is free from the restriction of editing relationship and the risk of out-of-distribution artifacts. The reconstruction improvement can be demonstrated both qualitatively and quantitatively in Fig.~\ref{fig:detok_compare}

\begin{figure}[t]
  \centering
   \includegraphics[width=0.99\linewidth]{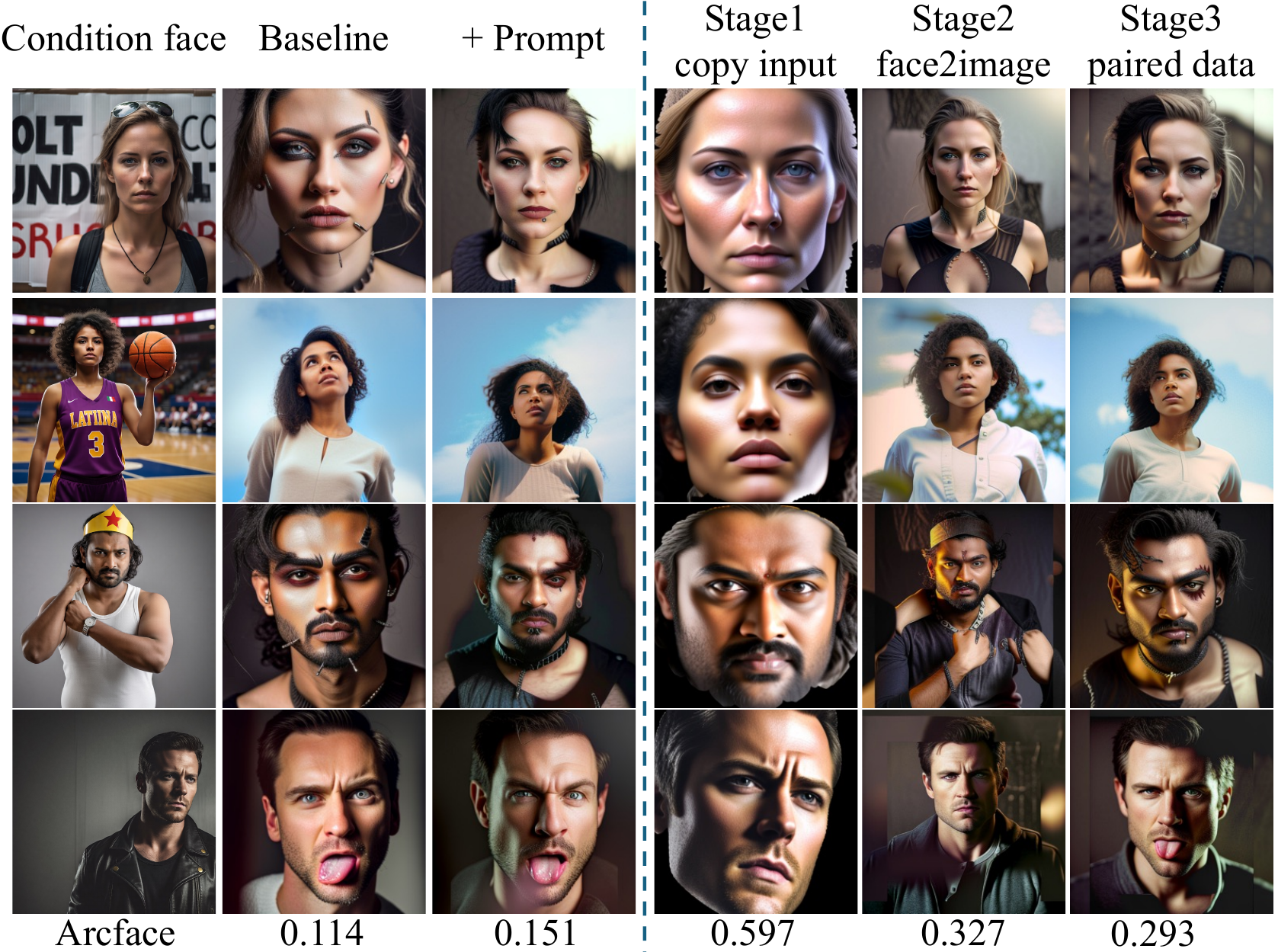}
   \caption{\textbf{Personalization results w.r.t. improvements.} See Appendix for prompts. Prompt instructions enhance face preservation slightly. Stage1 yields identical face generation; Stage2 maintains face similarity with editability beyond faces but struggles with complex prompts which request face modification; Stage3 achieves the best balance between face identity and editability.}
   \label{fig:multistage}
\end{figure}

\subsection{Enabling Component II: Multi-Stage Instruction Fine-Tuning for Editable Personalization}
\label{sec:multistage_ift}

Even with a strong detokenizer, a generative MLLM does not automatically preserve identity in conditional generation: As shown in Fig.~\ref{fig:multistage}, naive instruction tuning can follow prompts while producing a different face. We therefore introduce a multi-stage instruction fine-tuning strategy that progressively transitions the model from face copying to editable personalization.

\textit{Prompt design for identity preservation:} We leverage the language understanding ability of MLLMs by adding explicit identity constraints to the instruction template during fine-tuning (e.g., ``Please keep the face identical''), optionally paired with a short confirmatory response (e.g., ``I keep the face unchanged'') before generating image tokens. This encourages the model to pay more attention on face and identity.

\subsubsection{Stage 1: identity-first warm-up}
We first train the model to generate an image that closely resembles the conditioning image, prioritizing identity fidelity even if prompt compliance is weak. This stage establishes a strong identity prior and prevents early collapse to ``prompt-only'' generation.

\subsubsection{Stage 2: region-focused conditioning}
Next, we condition on masked face regions and train the model to reconstruct the full image. This teaches the model to preserve identity-critical facial regions while allowing edits elsewhere, improving the identity--editability trade-off.

\subsubsection{Stage 3: paired-identity supervision}
Finally, we fine-tune using paired images of the same person (different poses, lighting, contexts). The model is trained to predict one image conditioned on the other person's face, enabling controlled edits and variations while maintaining identity. To avoid overfitting and preserve robustness, we regularize this stage with a portion of Stage-2 data.

In summary, Stage-1 prevents identity collapse; Stage-2 adds ability of prompt following and background editing; Stage-3 provides direct supervision for editable identity preservation under real variations. As demonstrated in Fig.~\ref{fig:multistage}, this progression yields a model that preserves the subject while still responding to complex prompts, including those requiring facial attribute changes.


\section{Experiments}
\label{sec:experiments}

\begin{figure}[t]
    \centering
    \includegraphics[width=0.99\linewidth]{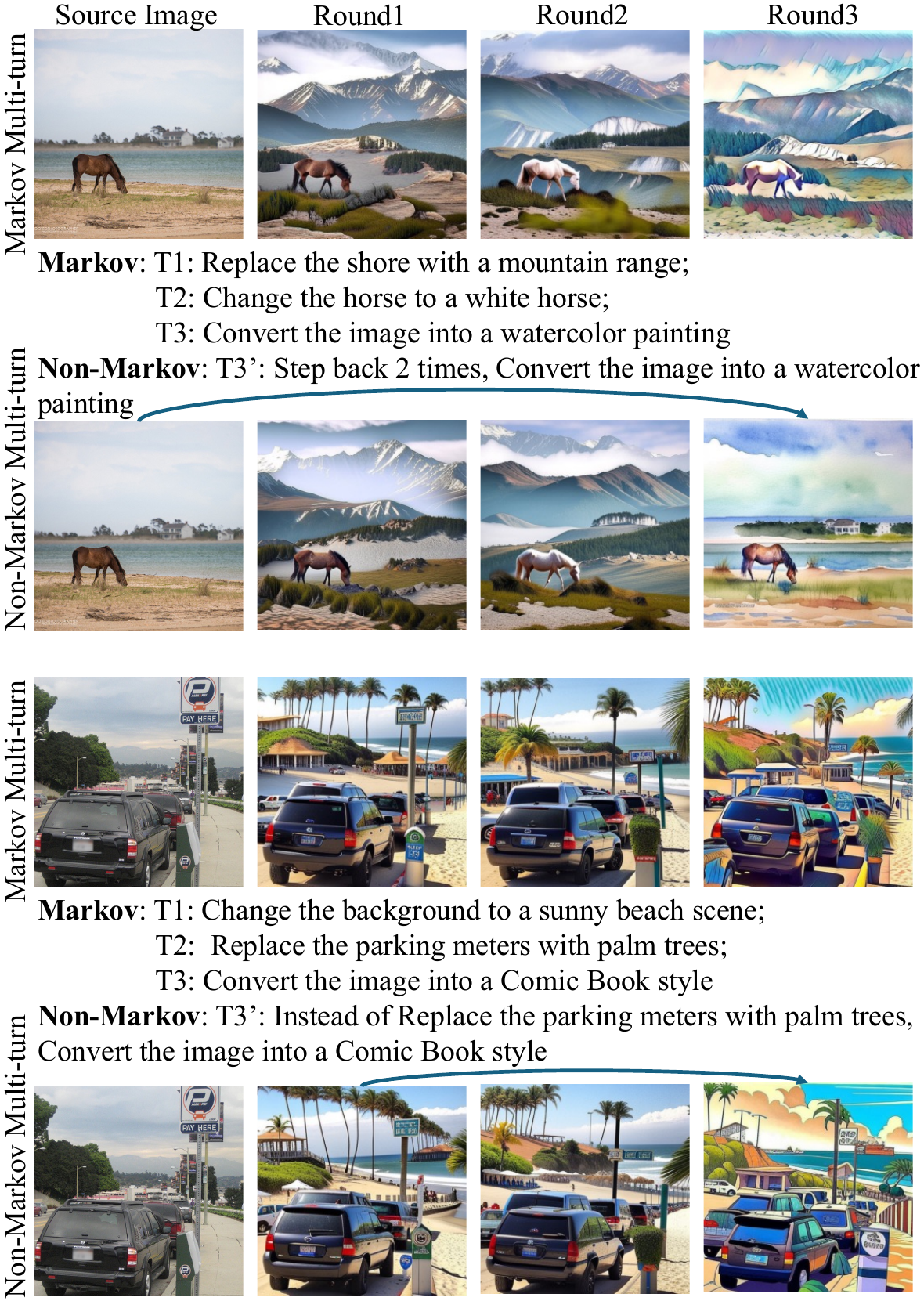}
    \caption{\textbf{Multi-turn editing results.} Our model handles Markov editing and accurately follows non-Markov rollback instructions by selecting the correct source image from history.}
    \label{fig:multiturn_edit}
\end{figure}

This section evaluates our model on two multi-round settings that explicitly require \emph{history reasoning}: (i) \textbf{non-Markov rollback-style editing}, where the editing source is not necessarily the most recent image, and (ii) \textbf{name-based multi-round personalization}, where later prompts may only reference a symbolic name and the requested identity may not appear in the immediately preceding output. We additionally include single-turn editing/personalization results to isolate the effects of our key components, detokenizer and instruction tuning, as ablations and analysis.

\begin{figure*}[t]
  \centering
   \includegraphics[width=0.99\linewidth]{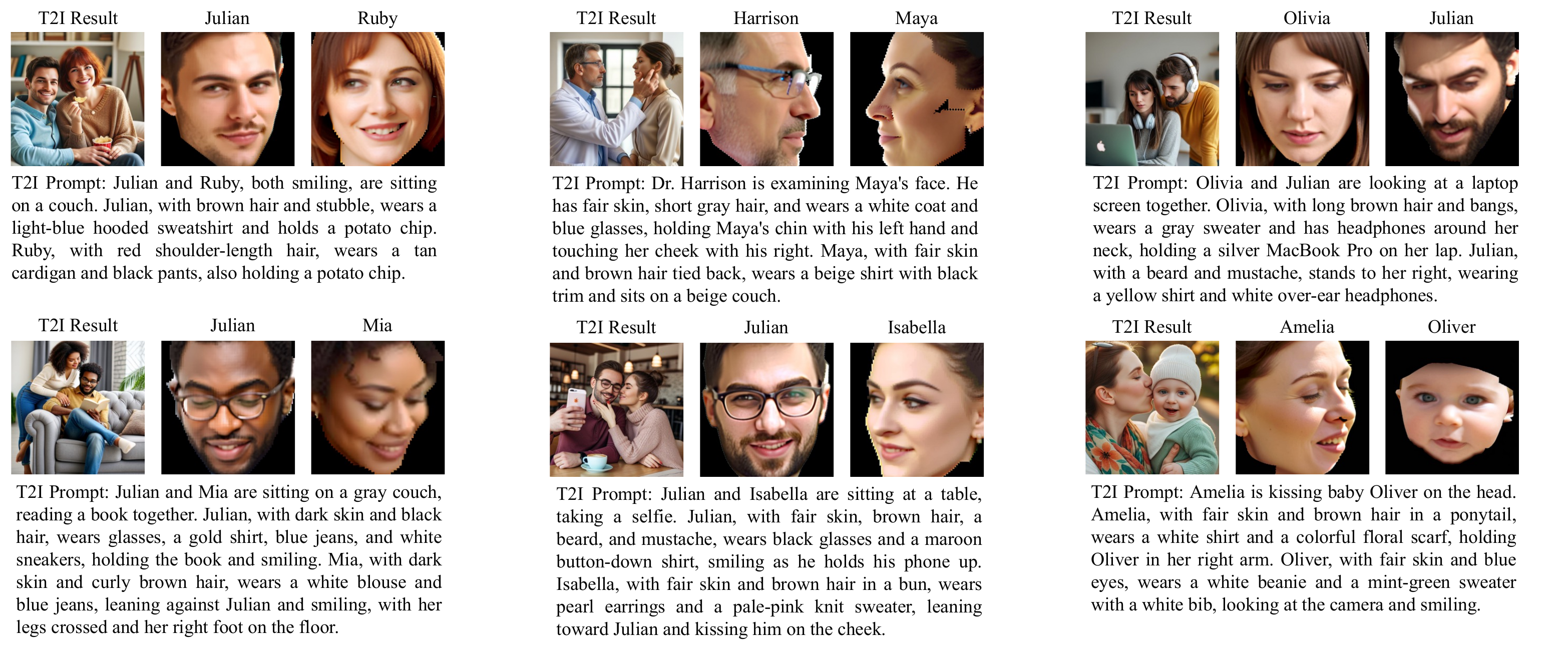}
   \caption{\textbf{Multi-turn personalization results.} The inference involves 3 rounds: Round 1 takes T2I prompt as input to generate an image; Round 2, 3 use prompt ``Generate a close-up photo of \{name\}" to generate faces. Our model generates images of the two individuals in the 1st round as well as faces of the correct individuals in personalization rounds, demonstrating its ability of retrieving earlier name--identity bindings.}
   \label{fig:multiturn_personal}
\end{figure*}

\begin{figure*}[t]
  \centering
   \includegraphics[width=0.99\linewidth]{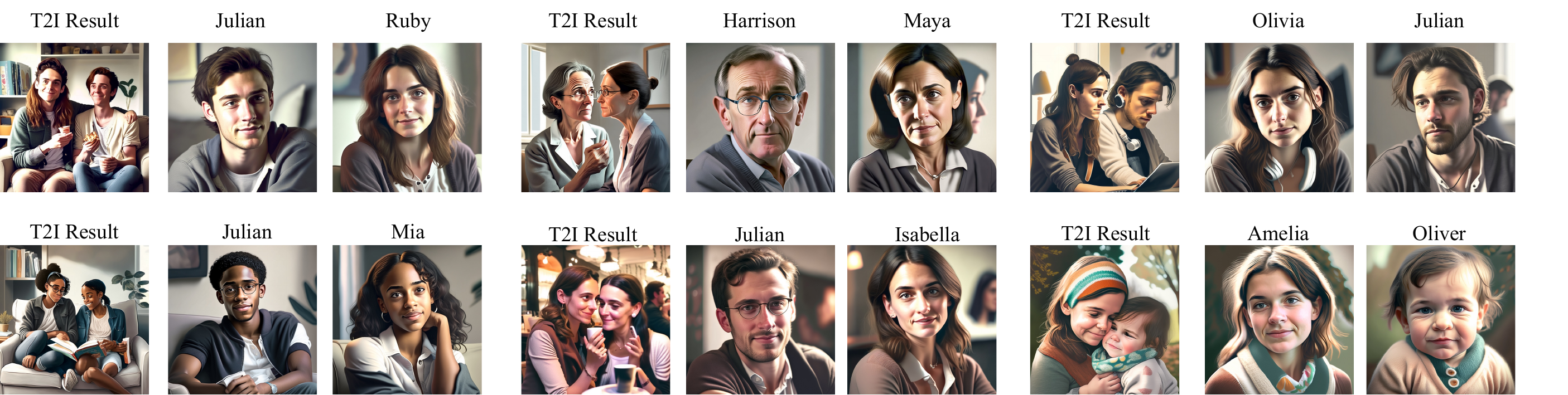}
   \caption{\textbf{Multi-turn personalization results of SEED-X} without fine-tuning on our proposed multi-turn dataset. These results can be directly compared to Figure~\ref{fig:multiturn_personal}. The baseline model struggles to generate reasonable two-person images in the first round and, in subsequent personalization rounds, behaves like a standard T2I model—regenerating faces from text cues while ignoring conditions established in the initial output.}
   \label{fig:s_seedx_multi}
\end{figure*}

\subsection{Implementation Details}

\subsubsection{Dataset Details} \label{sec:exp_dataset}
Here, we report key statistics for our datasets constructed in Sec.~\ref{sec:rollback_data} and Sec.~\ref{sec:name_data}.

\textit{Rollback-style multi-turn editing.}
We build on SEED-Data-Edit~\cite{ge2024seed} and retain samples with $\ge$4 editing rounds (for longer samples, we keep the first four rounds). Following Fig.~\ref{fig:rollback}, we generate six additional single-turn edit results per Markov chain using Emu Edit~\cite{sheynin2024emu}, and rearrange them into six rollback non-Markov multi-turn samples. Starting from 13,041 Markov chains (id$\le$20799), we obtain 78,246 Emu-Edit generations. We then filter noisy targets using CLIP-score-based screening, resulting in 54,261 non-Markov multi-turn training samples.

\textit{Name-based multi-turn personalization.}
We start from $\sim$170K two-person video clips and filter out those where the subjects are too far apart to fit within a $512\times512$ box, leaving $\sim$150K videos. We use LLaMA-3~\cite{dubey2024llama} to caption frames and rewrite captions with explicit name bindings, and keep samples containing exactly two names, yielding 92,471 training samples. For the optional \textbf{full-body} variant (Sec.~\ref{sec:exp_fullbody}), we filter diffusion-based personalization outputs using ArcFace~\cite{deng2019arcface} similarity, obtaining 24,793 full-body training samples.

\subsubsection{Training Recipes} \label{sec:exp_recipes}
This section summarizes all training details about how we finetune our MLLM framework as well as necessary enabling components.

\textit{DiT detokenizer.}
We fine-tune a DiT detokenizer using the same dataset and protocol as~\cite{polyak2024movie}. An MLP adapter is added on top of DiT to match the Qwen-VL~\cite{bai2023qwen} feature dimension. We use a constant learning rate of $10^{-5}$ with an effective batch size of 1024 (smaller batches do not converge). Training runs for 180K iterations on nature images and 96K iterations on human images.

\textit{LLM fine-tuning with LoRA.}
For LLaMA fine-tuning, we generally adhere to the default settings of SEED-X, incorporating necessary modifications. We initialize from a pretrained LlamaForCausalLM and fine-tune with LoRA~\cite{hulora}. We use AdamW ($\beta_1{=}0.9$, $\beta_2{=}0.98$, $\epsilon{=}10^{-6}$), weight decay 0.05, max grad norm 1.0, and mixed precision (\texttt{bf16}). Unless noted, we train for 60K iterations on a single node with 8 GPUs and use gradient accumulation to reach the desired effective batch size.

\textit{Single- and multi-turn editing fine-tuning.}
Single-turn editing is fine-tuned from SEED-X using a cosine schedule from $10^{-4}$ to $5\times10^{-6}$, with effective batch size 1024 and LoRA rank/$\alpha$ of 256. We sample Emu Edit~\cite{sheynin2024emu} and SciQA~\cite{auer2023sciqa} data with equal probability. Multi-turn editing then continues from the single-turn model using a step lr schedule ($10^{-6}\rightarrow5\times10^{-7}$ at 30K iterations). Data are sampled with ratio non-Markov : Markov : Emu Edit : SciQA $=4:2:1:1$, and other settings match single-turn training.

\textit{Single- and multi-turn personalization fine-tuning.}
We set LoRA rank/$\alpha$ to 1280 for all stages. Stage~1 trains identity-first copying with learning rate $10^{-5}$, effective batch size 1024, for 6K iterations. Stage~2 conditions on a cropped face plus caption to predict the full image with learning rate $10^{-6}$, effective batch size 512, for 30K iterations. Stage~3 introduces paired same-identity supervision with learning rate $10^{-7}$, effective batch size 1024, for 24K iterations, mixing Stage~2 and Stage~3 data with a fixed 1:2 ratio. For multi-turn, we train with a mixed dataset: multi-turn name-based data (Sec.~\ref{sec:name_data}), plus two augmentations derived from Stage~2/3 data (Agmnt1: full-image prediction for T2I; Agmnt2: face prediction for personalization), and SciQA regularization. The mixture ratio is multi-turn : Agmnt1 : Agmnt2 : SciQA $=6:2:3:1$. We use learning rates $10^{-4}$, $10^{-5}$, and $10^{-6}$ for 28K, 50K, and 12K iterations, respectively, with effective batch size 512.

\begin{figure*}[t]
    \centering
    \includegraphics[width=0.99\linewidth]{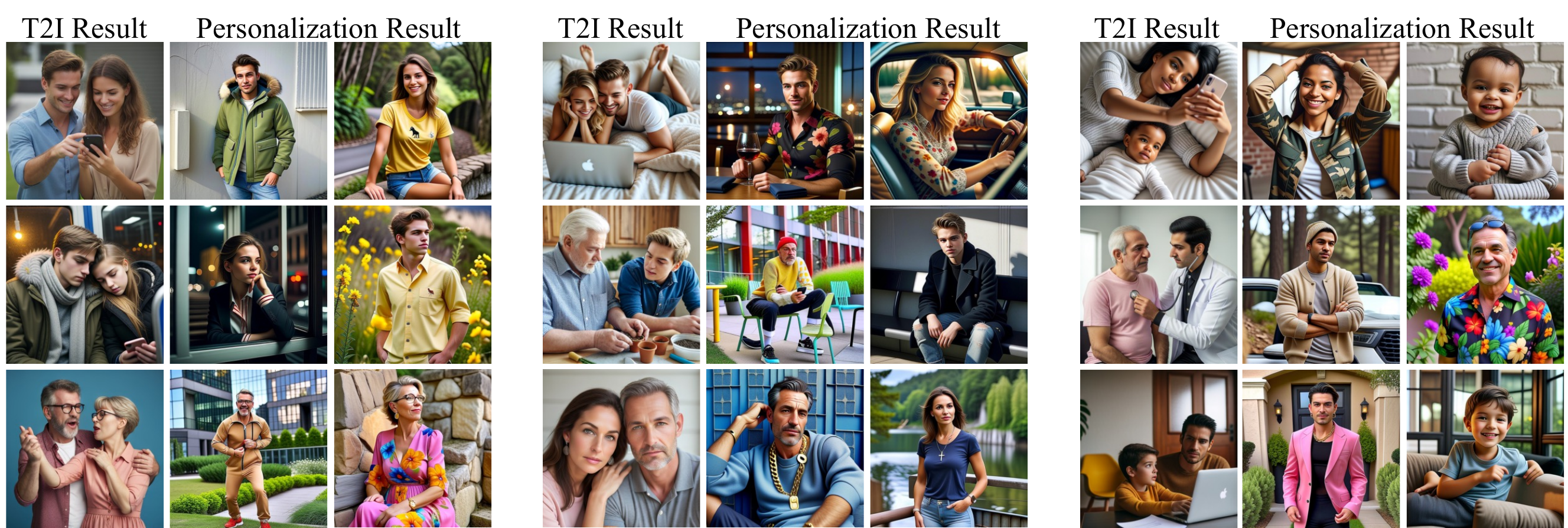}
    \caption{\textbf{Full-body multi-turn personalization results.} Fine-tuning on this dataset enables consistent full-body identity across multiple rounds. While the model follows detailed full-body prompts well, face fidelity is lower than in close-up portraits (Fig.~\ref{fig:multiturn_personal}) due to imperfect identity preservation in synthesized full-body supervision and the smaller size of the full-body dataset.}
    \label{fig:fullbody_personal}
\end{figure*}

\subsection{Non-Markov Multi-Round History Reasoning}
\label{sec:exp_main}

Due to the novelty of the multi-round image generation task, there is limited prior work for direct comparison. Thus, we evaluate the effectiveness mainly through visual inspection. 

\subsubsection{Non-Markov rollback editing}
Fig.~\ref{fig:multiturn_edit} shows that our model performs standard Markov multi-turn editing and, crucially, can follow rollback instructions that require selecting an earlier image in the history as the editing source (non-Markov).
This directly tests whether the model can interpret text-image interleaved history, identify the correct ``state'' to roll back to, and apply the new edit without being distracted by intermediate turns.

In the Appendix, we present additional multi-turn editing results for our model (Fig.~\ref{fig:multiturn_edit_res_supp}) and for the SEED-X-Edit model equipped with our DiT detokenizer (Fig.~\ref{fig:multiturn_edit_res_seedx_supp}). These results show that, even with an improved detokenizer, models that are not instruction-tuned on the proposed rollback-style non-Markov dataset consistently fail to follow rollback instructions and select the correct historical source image for editing.

\subsubsection{Name-based multi-round personalization} In this setting, later user prompts may contain only a name (e.g., ``Generate a close-up photo of Naomi''), and the requested identity in Round~3 may not be present in the Round~2 output. The model must retrieve the name--appearance binding introduced in Round~1 by jointly reasoning over the first-round text and the first-round generated image tokens. 

We present representative multi-round generations in Fig.~\ref{fig:multiturn_personal}. As observed, our model effectively generates images of the two individuals in the first round, despite minor mismatches, such as “red shoulder-length hair.” More importantly, in the second and third rounds, our model generates faces of the correct individuals as appeared in the first round output. Additional examples are provided in Appendix Fig.~\ref{fig:multiturn_person_res_supp}.

Fig.~\ref{fig:s_seedx_multi} reports baseline SEED-X multi-turn personalization results. As can be observed, firstly, this baseline model has difficuties in generate a reasonable two-person images in the first round; Then in personalization rounds, this model functioned similarly to a T2I model by identifying text near the name and generating a new face using its T2I capability, entirely disregarding the first-round output conditions.

\subsubsection{Name-Based Full-Body Personalization Extension}\label{sec:exp_fullbody}
Beyond face-centric portraits, we extend name-based multi-round personalization to \textbf{full-body} generation. Starting from the segmented faces in the name-based dataset (Fig.~\ref{fig:name_dataset}), we pair full-body personalization prompts with a diffusion-based personalization model~\cite{he2024imagine} to synthesize full-body supervision. This yields a 3-round structure: \textbf{Round~1}: T2I generation introducing and naming two individuals, and \textbf{Round~2/3}: name-based full-body personalization for each individual exampled as in Fig.~\ref{fig:fullbody_personal}.

\begin{figure*}[t]
  \centering
   \includegraphics[width=0.99\linewidth]{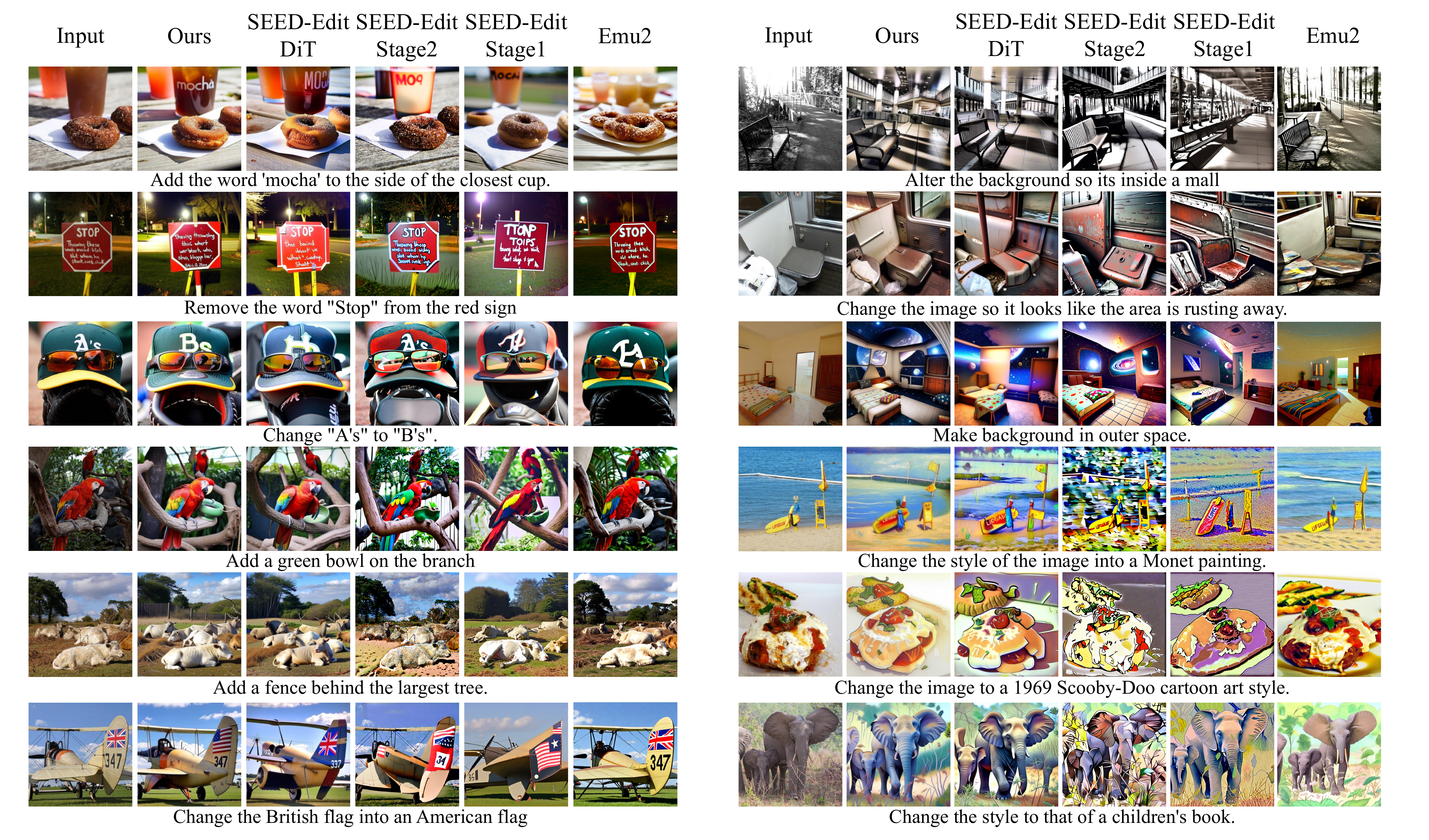}
   \caption{\textbf{Detokenizer ablation on single-turn editing.} Stage-1 can lose fine details; Stage-2 can introduce artifacts when the condition is out-of-distribution; Our DiT excels in content preservation among all detokenizers. When combined with our finetuned LLaMA model, Ours column demonstrates superior instruction-following ability.}
   \label{fig:singleturn_edit}
\end{figure*}

Fine-tuning on this dataset enables multi-round full-body personalization. While the model generally follows detailed full-body prompts, face fidelity is weaker than in the close-up portrait setting (Fig.~\ref{fig:multiturn_personal}). This is expected: full-body supervision is synthesized via diffusion models~\cite{he2024imagine}, and despite ArcFace-based filtering, identity preservation in the generated ground truth is imperfect. In addition, the full-body dataset is substantially smaller than the face-centric one (92K v.s. 24K detailed in Sec.~\ref{sec:exp_dataset}), making the task inherently more challenging.

To further probe history usage in the full-body setting, we repeat inference with identical text prompts like Appendix Fig.~\ref{fig:multiturn_person_res_memu_supp}. Because Round~1 does not explicitly specify attributes such as age, the model may generate different plausible identities (e.g., a child versus a teenager). Importantly, subsequent rounds consistently preserve these inferred attributes, indicating that the model retrieves identity information from earlier conversational image context rather than relying solely on textual prompt. Take the second result as an example: if `Lucas' appears as a teenager in the first-round T2I result, he should continue to appear as a teenager in the third personalization result. Thus, despite remaining challenges in face fidelity, these results support our central claim that the proposed framework enables non-markov conversational reasoning and consistent identity maintenance across multiple rounds.

\subsection{Ablations on Single-Turn Editing and Personalization}
\label{sec:ablation_singleturn}

Although our main contribution targets non-Markov multi-round generation, two enabling components are essential for making the framework work in practice: (i) a high-fidelity detokenizer (Sec.~\ref{sec:dit_detok}) and (ii) a multi-stage fine-tuning strategy for editable personalization (Sec.~\ref{sec:multistage_ift}). For simplicity, we do ablation experiments for these two components based on single-turn editing and personalization settings: single-turn editing isolates the impact of the \textbf{detokenizer}, while single-turn personalization isolates the impact of \textbf{multi-stage fine-tuning}.

\begin{figure*}[t]
  \centering
   \includegraphics[width=0.98\linewidth]{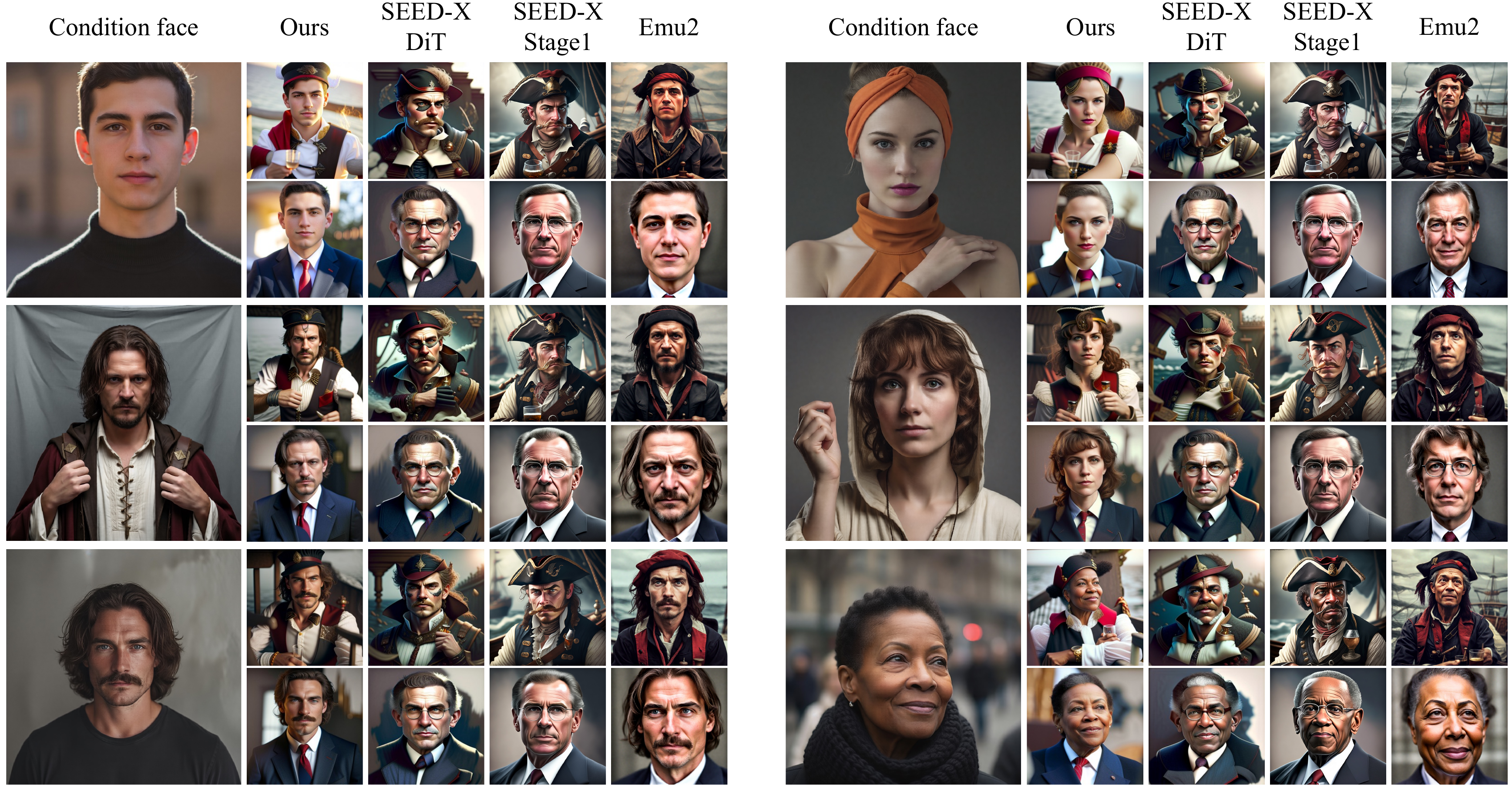}
   \caption{\textbf{Multi-stage fine-tuning ablation on single-turn personalization.} SEED-X and EMU2 struggle to perserve human faces especially when the condition images and the prompts have conflicts. I.e. they prioritize prompt semantics over visual conditioning. Our model shows the best tradeoff between face perseveration and prompt alignment. More visual inspection can be found in Figure~\ref{fig:singleturn_person_supp}.}
   \label{fig:singleturn_person}
\end{figure*}

\subsubsection{Detokenizer Ablation via Single-Turn Editing}
\label{sec:ablation_detok_edit}
We mainly compare three decoding choices used throughout the drafts: (i) \textbf{SDXL Stage-1} reconstruction detokenizer; (ii) \textbf{SDXL Stage-2} condition-image based detokenizer; and (iii) our \textbf{reconstruction-based DiT detokenizer}. The key distinction is that Stage-2 relies on a condition-image pathway tuned for editing-like relationships, while our DiT detokenizer decodes from image tokens alone and remains agnostic to whether the input arises from editing or personalization.

\begin{table}[t]
\caption{Human evaluation results for single-round editing.\label{tab:human_edit}}
\centering
\begin{tabular}{|c||c|c|c|}
\hline
\textbf{Metric} & \textbf{Ours Wins} & \textbf{SEED-X-Edit Wins} & \textbf{Tie} \\
\hline
Image Quality & 58\% & 2\% & 40\% \\
\hline
Prompt Alignment & 39\% & 9\% & 52\% \\
\hline
Content Preservation & 69\% & 2\% & 29\% \\
\hline
\end{tabular}
\end{table}

The three central columns of Fig.~\ref{fig:singleturn_edit} show our qualitative evidence. On single-turn editing, Stage-1 tends to preserve coarse semantics but can distort fine structures, which harms content preservation. Stage-2 improves preservation when the condition-target relationship matches its training distribution, but can introduce visible artifacts or unnatural textures when the condition is out-of-distribution. In contrast, our DiT detokenizer consistently preserves fine details without requiring an extra condition input, yielding more stable editing outputs across diverse prompts. We also calculated PSNR between input and reconstructed image on a subset of COCO2014~\cite{lin2014microsoft} images and report numbers in Fig.~\ref{fig:detok_compare}, which demonstrates a significant improvement in reconstruction quality with our DiT compared to the baseline SDXL detokenizer.

Comparing the editing results, EMU2 often ignores the editing instruction and outputs analogues of input. SEED-X-Edit, with its LLaMA model finetuned on editing data, performs editing tasks but often fails to complete tasks well, even when equipped with the DiT detokenizer. Conversely, our model excels across all tasks. We illustrate more editing results from our model in Appendix Fig.~\ref{fig:singleturn_edit_supp}. In the single-turn editing user study summarized Table~\ref{tab:human_edit}, we manually evaluated 400 samples randomly selected from Emu Edit test set~\cite{sheynin2024emu}. As can be observed, our method is preferred more often on content preservation, while remaining competitive on image quality and prompt alignment. The gains can be attributed to the improved decoding fidelity of our DiT detokenizer.

\begin{table}[t]
\caption{Human evaluation results for single-round personalization.\label{tab:he_person}}
\centering
\begin{tabular}{|c||c|c|c|}
\hline
\textbf{Metric} & \textbf{Ours Wins} & \textbf{SEED-X Wins} & \textbf{Tie} \\
\hline
Image Quality & 73.75\% & 3.75\% & 22.5\% \\
\hline
Prompt Alignment & 36.25\% & 15\% & 48.75\% \\
\hline
Face Identity Preservation & 71.25\% & 2.5\% & 26.25\% \\
\hline
\end{tabular}
\end{table}

\subsubsection{Multi-Stage Fine-Tuning Ablation via Single-Turn Personalization}
\label{sec:ablation_multistage_personal}
Name-based multi-round personalization relies on retrieving identity bindings, but it also requires a simpler prerequisite: given a conditioning face, the model must preserve identity while following prompts that change pose and expression. Single-turn personalization therefore serves as a clean ablation for multi-stage fine-tuning, directly measuring the identity–editability trade-off without confounds from multi-round name-appearance binding.

Fig.~\ref{fig:singleturn_person} compares single-turn personalization results from SEED-X, EMU2, and our model (see Appendix for prompts). Since SEED-X does not release an instruction-finetuned personalization model, we compare against its pretrained version. As shown, SEED-X struggles to preserve facial identity regardless of the detokenizer, behaving largely as a text-to-image model. EMU2 partially preserves identity in simpler cases (left examples); however, it fails when the prompt conflicts with the conditioning image. For instance, prompts such as `pirate captain' or `USA president look' introduce strong gender priors, causing EMU2 to override the conditioning face and generate incorrect identities (right examples). In contrast, our model consistently preserves identity while remaining aligned with the requested attributes. Additional qualitative results are provided in Appendix Fig.~\ref{fig:singleturn_person_supp}.

We further support these observations with quantitative evidence. During development, we track ArcFace~\cite{deng2019arcface} and CLIP~\cite{ruiz2023dreambooth} scores as diagnostic metrics. Our model shows a substantial improvement in ArcFace similarity (0.094 $\rightarrow$ 0.293) while maintaining comparable CLIP scores (28.36 $\rightarrow$ 28.59), indicating stronger identity preservation without sacrificing semantic alignment. Final human evaluation (Table~\ref{tab:he_person}) on a 400-subset of~\cite{he2024imagine} test set confirms these gains: our model consistently outperforms SEED-X across image quality and face preservation. Notably, SEED-X ties with ours in prompt alignment because it behaves like a text-to-image model, often ignoring the conditioning face, as illustrated by the visually similar `USA president look' outputs despite different conditioned identities in Fig.~\ref{fig:singleturn_person}.

Since our model is fine-tuned from SEED-X, these improvements directly demonstrate the effectiveness of the proposed multi-stage fine-tuning strategy in enforcing identity preservation while retaining prompt responsiveness.


\section{Conclusion}
\label{sec:conclusion}

In this work, we investigated \textbf{non-Markov multi-round conversational image generation}, a more general setting of text--image interleaved generation that more faithfully reflects how users interact with generative models. We showed that many existing multi-round approaches rely on Markov shortcuts and fail when later instructions depend on earlier conversational states or long-range symbolic references. To address this gap, we introduced two non-Markov dataset constructions—rollback-style multi-round editing and name-based multi-round personalization—and a history-conditioned training and inference framework compatible with the SEED-X generation interface. Together with token-level history caching, a reconstruction-based DiT detokenizer, and a multi-stage personalization fine-tuning strategy, our approach enables reliable long-horizon reasoning while remaining strong on single-round editing and personalization.

Our study also highlights several limitations and directions for future work. First, the lack of standardized benchmarks for personalization and non-Markov multi-round generation makes evaluation largely qualitative and human-driven; developing shared benchmarks would greatly benefit the field. Second, our current name-based dataset primarily uses face-centric supervision for later rounds; extending this to full-body and multi-object personalization remains an open challenge. Finally, while we focus on rollback and name-based references, real conversations involve even richer long-range dependencies (e.g., relational references and compositional memory), which we view as promising directions for future conversational image generation research.





{
\appendix

\begin{figure*}[t]
  \centering
   \includegraphics[width=0.99\linewidth]{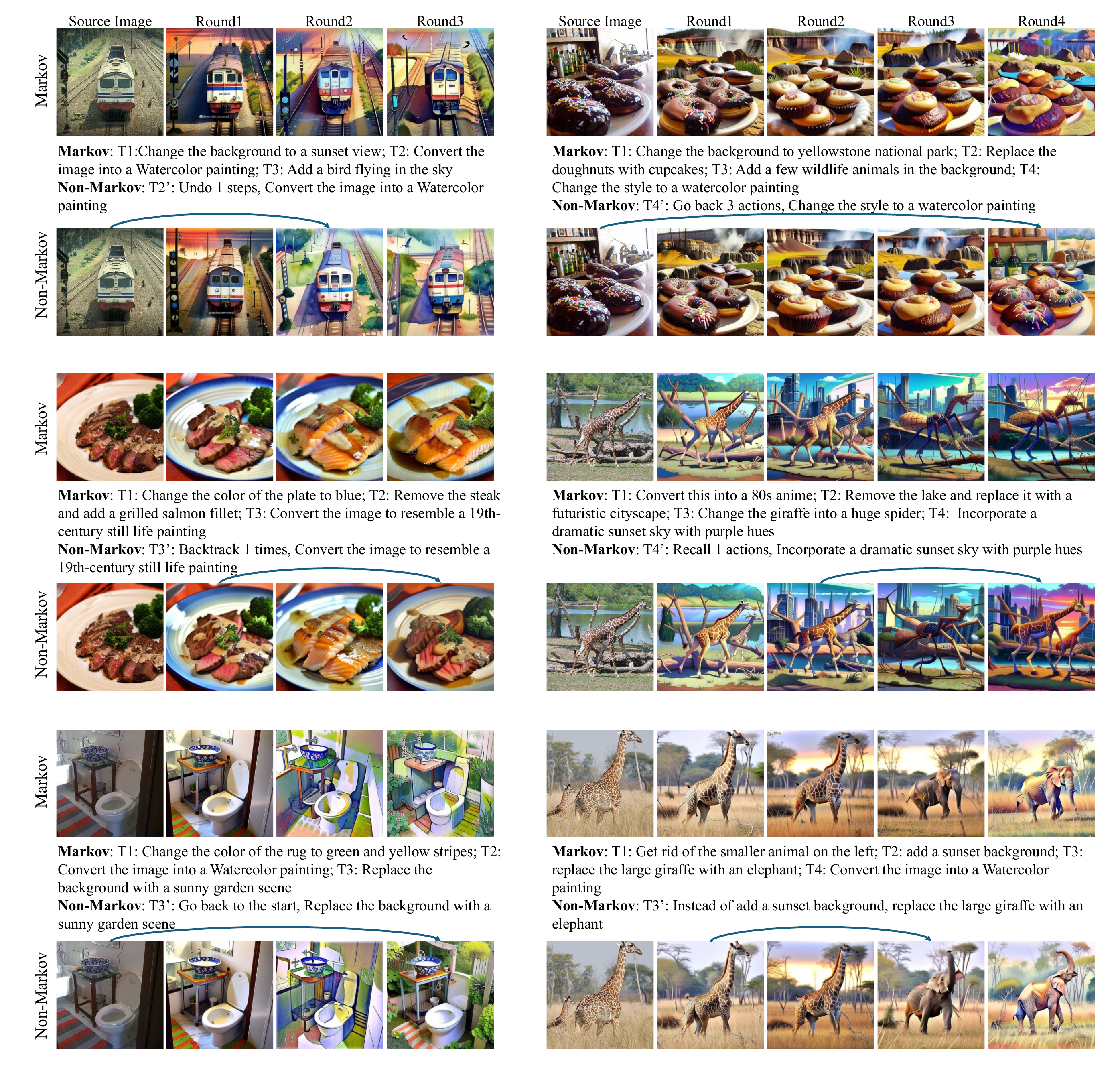}
   \caption{\textbf{More multi-turn editing results from our MLLM in Markov and non-Markov settings.} Please compare with Fig.~\ref{fig:multiturn_edit_res_seedx_supp}.}
   \label{fig:multiturn_edit_res_supp}
\end{figure*}

\begin{figure*}[t]
  \centering
   \includegraphics[width=0.99\linewidth]{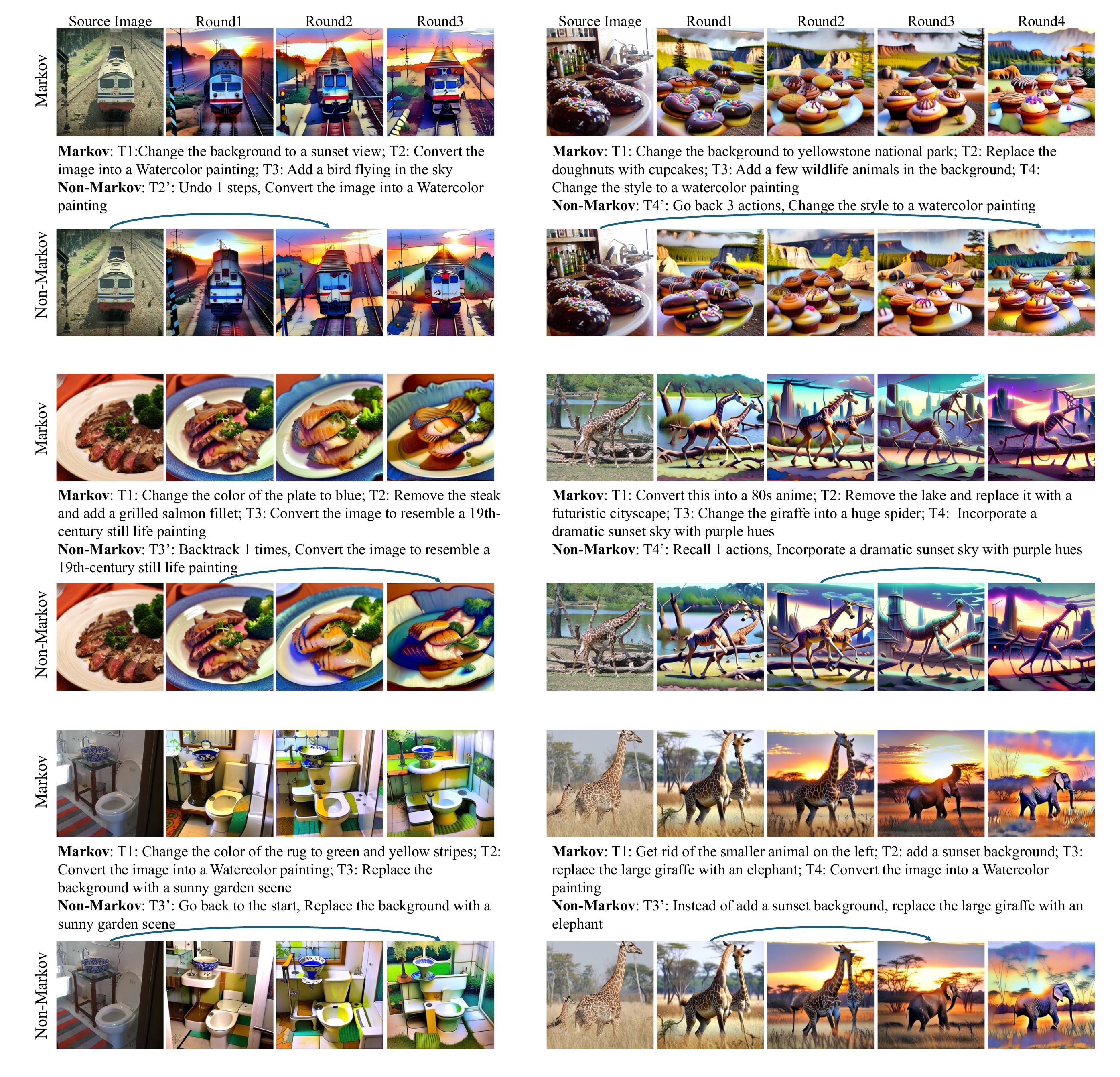}
   \caption{\textbf{Multi-turn editing results from SEED-X-Edit model} equipped with our DiT detokenizer. The performance drop compared with Figure~\ref{fig:multiturn_edit_res_supp} demonstrates the advantages of the LLaMA component finetuned on our proposed training data and strategy.}
   \label{fig:multiturn_edit_res_seedx_supp}
\end{figure*}

\begin{figure*}[t]
  \centering
   \includegraphics[width=0.99\linewidth]{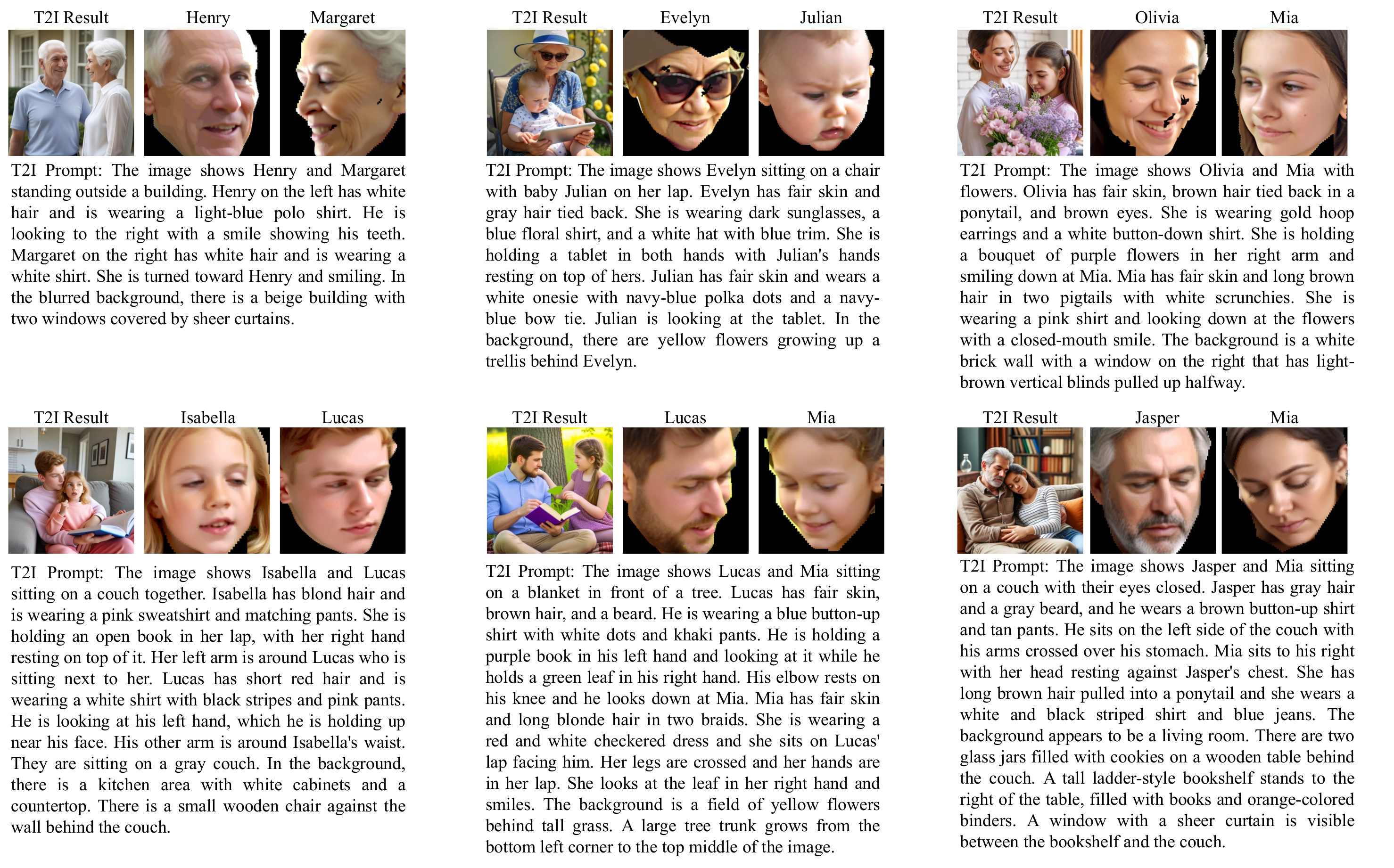}
   \caption{\textbf{More multi-turn personalization results.}}
   \label{fig:multiturn_person_res_supp}
\end{figure*}

\begin{figure*}[t]
  \centering
   \includegraphics[width=0.99\linewidth]{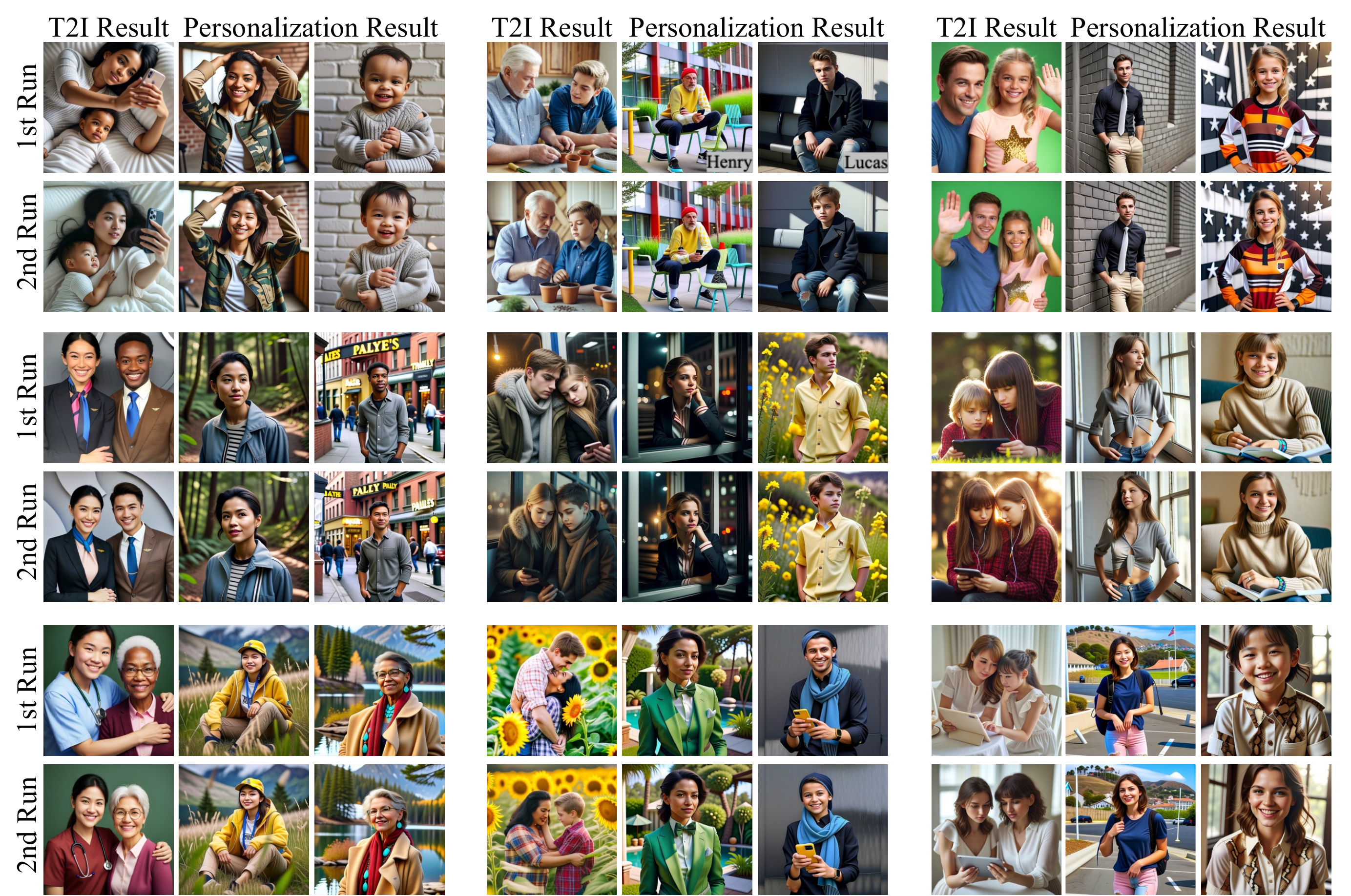}
   \caption{\textbf{Further explorations on full-body multi-turn personalization.} As can be observed, when inferring twice with the same prompt, distinct 1st-turn T2I results are generated. Subsequently, the 2nd- and 3rd-turn personalization results are different as well. For instance, if the model initially generates an image containing a boy rather than a teenager, the subsequent personalization results will also depict a boy. This behavior is a strong evidence that our model can generate image based on reasoning from text-image interleaved chat histories.}
   \label{fig:multiturn_person_res_memu_supp}
\end{figure*}

\begin{figure*}[t]
  \centering
   \includegraphics[width=0.99\linewidth]{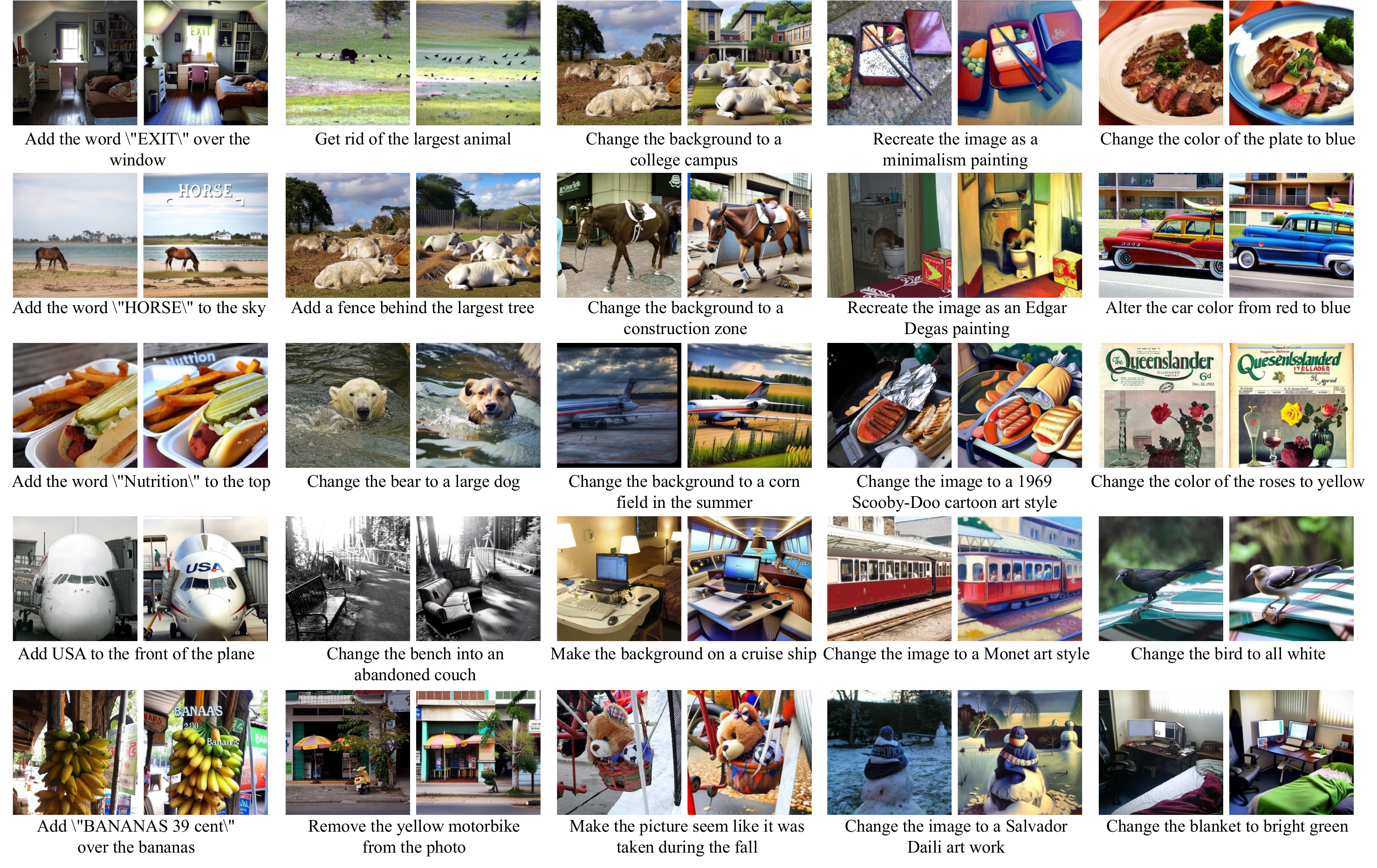}
   \caption{\textbf{More single-turn editing results.}}
   \label{fig:singleturn_edit_supp}
\end{figure*}

\begin{figure*}[t]
  \centering
   \includegraphics[width=0.99\linewidth]{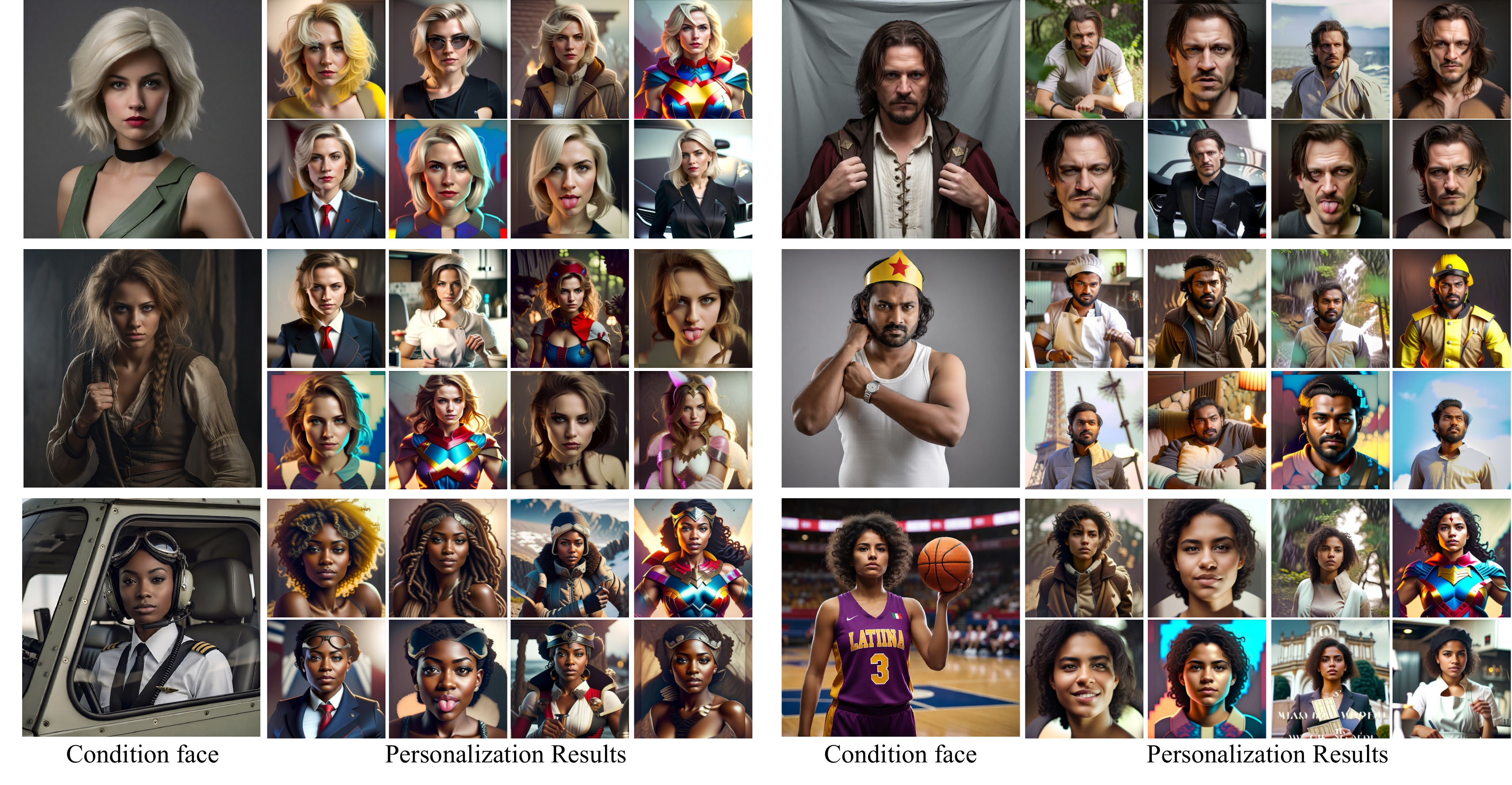}
   \caption{\textbf{Single-turn personalization visual examples of our MLLM.}}
   \label{fig:singleturn_person_supp}
\end{figure*}

\section*{Personalization Prompts}

The full prompts used in Figure~\ref{fig:multistage} are:

1\&3) ``A person with goth makeup, with their face visible and distinguishable. The person has a pale complexion, with dark eyeliner and mascara accentuating their eyes. Their lips are painted a deep red, and their eyebrows are plucked and drawn on to create a sharp, angular shape. A silver stud pierces their left eyebrow, and a choker made of black leather adorns their neck. The person's hair is black and styled in a messy, spiky fashion, adding to their goth aesthetic.''

2) ``A person looking up into the sky, with their face visible and distinguishable. The person is standing with their feet shoulder-width apart, their eyes squinting slightly as they gaze upwards. They are wearing a light-colored shirt and jeans, and their hair is blowing gently in the wind. The sky above is a brilliant blue, with only a few wispy clouds scattered across it. The person's expression is one of wonder and awe, as if they are marveling at the vastness of the sky.''

4) ``A person sticking out their tongue, with their face visible and distinguishable. The person's eyes are wide open, and their eyebrows are raised, creating a comical expression. Their tongue is bright pink and slightly curled, adding to the playful appearance. The person's face is positioned close to the camera, emphasizing the tongue-sticking-out gesture. The image is cropped closely around the person's face, focusing attention on the tongue and facial expression.''

~

\noindent The full prompts used in Figure~\ref{fig:multiturn_personal} are:

1) ``The image shows Julian and Ruby sitting on a couch together, smiling at the camera. Julian has fair skin with brown hair and stubble. He is wearing a light-blue hooded sweatshirt and holding a potato chip in his right hand. Ruby has fair skin with red shoulder-length hair. She is wearing a tan cardigan and black pants. She is holding a potato chip in her left hand. The background appears to be a living room. There are white shelves on the left with books and decorations on them. On the right, there is a tall floor lamp with a yellow lampshade and a white bookcase behind it." 

2) ``The image shows Dr. Harrison examining Maya's face. Dr. Harrison has fair skin and short gray hair, is wearing a white coat and blue glasses, and is holding Maya's chin with his left hand while he looks at her face with his right eye closed and his right hand touching her cheek. Maya has fair skin and brown hair tied back, is wearing a beige shirt with black trim and is sitting on a beige couch. The background is a room with white walls and black-framed windows." 

3) ``The image shows Olivia and Julian looking at a laptop screen together. Olivia, with long brown hair and bangs, looks down at a silver MacBook Pro that she holds on her lap. She is wearing a gray sweater and has headphones around her neck. Julian, with a beard and mustache, stands to her right, also looking down at the laptop screen. He is wearing a yellow shirt and white over-ear headphones. The background is a blurred room with white walls and a window on the left." 

4) ``The image shows Julian and Mia sitting on a gray couch, reading a book together. Julian has dark skin and black hair, and he's wearing glasses, a gold shirt, blue jeans, and white sneakers. He's holding an open book with both hands, looking down at it and smiling. Mia has dark skin and curly brown hair. She's wearing a white blouse and blue jeans, and she's leaning against Julian, looking down at the book and smiling. Her legs are crossed, and her right foot is resting on the floor. The couch is light-gray with two tufted seats and two matching throw pillows. Behind them, there's a tall, dark-gray metal bookshelf with a woven basket on top. A green plant peeks out from behind the basket. In front of the couch, there's a window with a white sheer curtain covering it."

5) ``The image shows Julian and Isabella sitting at a table in a restaurant, taking a selfie. Julian has fair skin and brown hair with a beard and mustache. He wears black glasses and a maroon button-down shirt. He sits on the left side of the table and smiles as he holds his phone up to take a selfie. Isabella has fair skin and brown hair pulled into a bun. She wears pearl earrings and a pale-pink knit sweater. She leans toward Julian and kisses him on the cheek while holding her hand under her chin. In front of them on the table are two white coffee cups and saucers. The background is blurred and appears to be a restaurant or cafe. There are hanging lights above the couple and more tables set with dishes and glassware behind them." 

6) ``The image shows Amelia kissing baby Oliver on the head. Amelia has fair skin and brown hair tied back in a ponytail. She is wearing a white shirt and a colorful scarf with orange, green, blue, black and yellow flowers on it. She is holding Oliver in her right arm who is looking at the camera and smiling. Oliver has fair skin and blue eyes. He is wearing a white beanie and a mint-green sweater with a white bib underneath. In the background there are trees and a path on the right side." 

~

\noindent The prompts used in Figure~\ref{fig:singleturn_person} are 

1) ``A person dressed as a pirate captain, with their face visible and distinguishable, standing at the helm of a ship navigating through the rough waters of the North Sea. The captain is wearing a white shirt with billowy sleeves, a red vest, and a black tricorn hat adorned with a golden chain and a feather. A whiskey tumbler glass is held tightly in their hand, with a hint of whiskey remaining at the bottom. The captain's facial expression is one of determination and focus, with a hint of ruggedness and weathered skin, suggesting a seasoned sailor." 

2) ``A person with a USA president look, with their face visible and distinguishable. They are wearing a navy blue suit with a white shirt and a red tie. A pair of glasses perches on the end of their nose, and a hint of a smile plays on their lips. Their hair is neatly combed and gray, suggesting a sense of wisdom and experience. The person exudes an air of confidence and authority, as if they are about to deliver an important speech or address the nation. The focus is on the person's face and upper body, with a blurred background that emphasizes their presence and leadership."

~

\noindent The prompts used in Figure~\ref{fig:multiturn_person_res_memu_supp} are relatively complex. Therefore, we present the second result (`Henry' and `Lucas') as a representative example.

\begin{itemize}
\item \textbf{Round1 text-to-image generation with two-person prompt including their names}: ``Generate a image shows \textit{Henry} and \textit{Lucas} sitting at a table together. Henry has white hair and a white beard, and he is wearing a blue-and-white checkered button-up shirt. He is looking down at his hands as he holds two small pots with brown dirt in them. There is a hand protruding from the bottom left corner of the image holding a handful of seeds that are spilling out into the pots. Lucas is on the right side of the table. He has blond hair and he is wearing a navy-blue button-up shirt. He is looking down at the table with a neutral expression. There are gardening tools on the table in front of him. The background shows a kitchen with light-brown wood panel walls. There is a white sink on the left edge of the image. Above it, there is a white countertop with a white faucet. On the back wall, there is a white electrical outlet with a white switch above it. There is a white cabinet underneath the countertop on the right side of the image".
\item \textbf{Round2 name-based personalization with full-body prompt}: ``\textit{Henry} is sitting on a light-green metal folding chair at an outdoor cafe. They wear a red beanie, a yellow long-sleeve shirt with a white checkered pattern, black pants, white socks, and black and white Nike shoes. Their left leg is bent upward and their right leg is stretched out behind them. They hold a phone in their left hand and look at the camera with a neutral expression. In front of them are two light-blue chairs and one light-yellow chair. The background is a gray sidewalk with green grass growing between it and a building. On the other side of the sidewalk is a glass wall with tall red spikes protruding from the ground."
\item \textbf{Round3 name-based personalization with full-body prompt}: ``\textit{Lucas} is sitting on a black bench. They are wearing a black peacoat over a black shirt and blue jeans with holes in the knees. They look at the camera with a neutral expression. The background is a white wall with a shadow falling on the left side."
\end{itemize}

\section*{Training Prompt Template}

For single turn personalization, we finetuned SEED-X on our personalization dataset using a prompt template: ``\texttt{<s> [INST] Generate the image shows \{caption\} <img> \{source embedding\} </img> [/INST] $\backslash$n I have generated an image. <img> \{target embedding\} </img> </s>}''. 

For single turn editing, we finetuned SEED-X on editing dataset using a prompt template: ``\texttt{<s> [INST] \{instruction\} <img> \{source embedding\} </img> [/INST] I have generated an image. \{target caption\} <img> \{target embedding\} </img> </s>}''. Here, we integrated the target image caption as additional text output to preserve the text generation ability.

For multi-round turn editing, we adopt a multi-round prompt template ``\texttt{<s>  [INST] \{1st round input (text+image)\} [/INST] $\backslash$n \{1st round response (text+image)\} $\backslash$n   [INST] \{2nd round input (text)\} [/INST] $\backslash$n \{2nd round response( text+image)\} $\backslash$n [INST] \{3rd round input (text)\} [/INST] $\backslash$n \{3rd round response (text+image)\} $\backslash$n </s>}".

\begin{table}[t]
\caption{Text and image scores of different models on SEED-Bench-2. Higher is better. Finetuning consistently decreases text score, while SciQA regularization preserves text performance without harming editing ability.}
\label{tab:textscore}
\centering
\begin{tabular}{|l||c|c|c|}
\hline
\textbf{Method} & \textbf{Text Score} & \textbf{Image Score} & \textbf{Total Score} \\
\hline\hline
SEED-X        & 0.899 & 0.734 & 0.684 \\
\hline
SEED-X-Edit    & 0.646 & 0.696 & 0.481 \\
\hline\hline
Ours-9K       & 0.772 & 0.696 & 0.595 \\
\hline
Ours-15K      & 0.506 & 0.722 & 0.380 \\
\hline
Ours-45K      & 0.392 & 0.722 & 0.266 \\
\hline\hline
Ours-60K      & 0.367 & 0.696 & 0.215 \\
\hline
\quad + SciQA & 0.899 & 0.658 & 0.608 \\
\hline
\end{tabular}
\end{table}

\section*{Editing Detail and Text Score}
In this section, we investigate the behavior of SEED-X in text generation and reasoning abilities when instruction finetuned with editing data. In detail, we finetuned SEED-X on editing dataset of ~\cite{sheynin2024emu} and we integrated the target image caption as additional text output, which is designed to preserve the text generation ability of SEED-X during the instruction fine-tuning process.

To assess the effectiveness of our finetuning approach, we utilize the ``Image \& Text Generation'' subset of SEED-Bench-2~\cite{li2024seed}. This evaluation subset presents a multi-choice question format, where the model must provide a text-based answer and subsequently generate an image that illustrates the answer. The model earns points for selecting the correct text answer and for generating the image with the highest CLIP similarity to the correct answer's corresponding image. Initial results, as reported in Table~\ref{tab:textscore}, indicate a consistent decrease in the text score of the finetuned model as training iterations increase, suggesting that merely predicting the target image caption is insufficient.

To address this problem, we introduce a stronger regularization by incorporating data from the SciQA dataset~\cite{auer2023sciqa}, mixed at a 1:1 ratio with the editing data. This regularization strategy not only maintains a high text score, as shown in Table~\ref{tab:textscore}, but surprisingly expedites the convergence of both text token cross-entropy loss and image token MSE loss as well. The improved editing performance of our finetuned model, facilitated by the SciQA regularization, has been detailed in Section~\ref{sec:exp_recipes}.

}

\bibliographystyle{IEEEtran}
\bibliography{main}

\vfill

\end{document}